\documentclass[final]{article}

\usepackage{neurips_2022}

\usepackage[utf8]{inputenc} 
\usepackage[T1]{fontenc}    
\usepackage{hyperref}       
\usepackage{url}            
\usepackage{booktabs}       
\usepackage{amsfonts}       
\usepackage{nicefrac}       
\usepackage{microtype}      
\usepackage{xcolor}         

\usepackage{tikz}
\usetikzlibrary{positioning}
\usetikzlibrary{calc,through,backgrounds}
\usepackage{pgfplots} 
\pgfplotsset{compat=newest}
\usepgfplotslibrary{groupplots}
\usepgfplotslibrary{dateplot}
\usepackage{algpseudocode}
\usepackage{algorithm}
\usepackage[export]{adjustbox}
\usepackage{amsmath,amsfonts,amssymb}
\usepackage{siunitx}
\usepackage{tabularx}
\usepackage{tabulary}
\usepackage{booktabs}
\usepackage{makecell}
\usepackage{multirow}
\usepackage{array}
\usepackage{colortbl}
\usepackage{dcolumn}
\usepackage{stfloats}
\usepackage{xspace,xstring,footmisc}
\usepackage{longtable}
\usepackage[export]{adjustbox}

\newcommand{\vc}[1]{\mathbf{#1}}     
\newcommand{\ma}[1]{\mathbf{#1}}     

\definecolor{mycolor}{cmyk}{1,0,1,0}
\definecolor{mycolor1}{rgb}{1.00000,0.50000,0.00000}%
\definecolor{mycolor2}{rgb}{0.00000,0.80000,1.00000}%
\definecolor{mycolor3}{rgb}{1.00000,0.00000,1.00000}%
\definecolor{mycolor4}{rgb}{0.45, 0.31, 0.59}%
\definecolor{mycolor5}{rgb}{0.6, 0.4, 0.8}
\definecolor{carnationpink}{rgb}{1.0, 0.65, 0.79}
\definecolor{auburn}{rgb}{0.43, 0.21, 0.1}

\newcommand{\new}[1]{{#1}}
\newcommand{\RED}[1]{{#1}}

\begin{document}
\let\WriteBookmarks\relax
\def\floatpagepagefraction{1}
\def\textpagefraction{.001}

\title{Crossing Points Detection in Plain Weave for Old Paintings with Deep Learning}

%

\author{%
 A.Delgado \\
  Dep. Teoría de la Señal y Comunicaciones.\\
   ETSI. Universidad de Sevilla.\\
  Camino de los Descubrimientos sn\\
  Sevilla, 41092. Spain 
   \And
   Laura~Alba-Carcelén. \\
   Dep. Restauración y Documentación Técnica
   Museo Nacional del Prado \\
   Paseo del Prado s/n \\
   28914 Madrid. Spain\\
   \AND
   Juan.~J.~Murillo-Fuentes\\
  Dep. Teoría de la Señal y Comunicaciones.\\
   ETSI. Universidad de Sevilla.\\
  Camino de los Descubrimientos sn\\
  Sevilla, 41092. Spain 
   \texttt{murillo@us.es} 
}

\maketitle

\begin{abstract}
In the forensic studies of painting masterpieces, the analysis of the support is of major importance. For plain weave fabrics, the densities of vertical and horizontal threads are used as main features, while angle deviations from the vertical and horizontal axis are also of help. These features can be studied locally through the canvas. 
In this work, deep learning is proposed as a tool to perform these local densities and angle studies. We trained the model with \new{samples} from 36 paintings by Velázquez, Rubens or Ribera, among others. The data preparation and augmentation \new{are} dealt with at a first stage of the pipeline. We then focus on the supervised segmentation of crossing points between threads. The U-Net with inception and Dice loss \new{are presented as good choices} for this task. Densities and angles are then estimated based on the segmented \new{crossing points}. We report test results of the analysis of a few canvases \RED{and a comparison with methods in the frequency domain, widely used in this problem. We concluded that this new approach} successes in some cases where the frequency analysis tools fail, while improves the results in \new{others}. Besides, our proposal does not need the labeling of part of the to be processed image. As \new{case studies}, we apply this novel \new{algorithm} to the analysis of \new{two pairs of canvases by Velázquez and Murillo}, to conclude that the fabrics used came from the same roll.

\end{abstract}



\new{
\section{Introduction}\label{sec:intro}
}

\subsection{Fabric Analysis and Plain Weave}


The weave and patterns on the fabric of paintings can be seen as features or fingerprints that help in the forensic analysis of paintings, e.g., to date and assign \new{authorship} 
\cite{Alba21}. The type of weave, the material, the number of threads per centimeter or the weight of fabrics have been widely studied \cite{Vanderlip80}. A photograph of the back of the painting can be processed to study the fabric, but in many cases the canvas has been relined to reinforce \new{it} 
and the original fabric cannot be observed directly. Instead, X-ray plates of the paintings are usually analyzed. Radiographs are much harder to process because the frame and the painting itself, including cracks, are observed as noise. 


Plain weave fabrics\footnote{Also known as tabby, calico or tafetta weave.} have been widely used as the support of paintings. In plain weave we have horizontal and vertical threads intertwined. In the loom, a set of threads, the warp, are parallel arranged from back to front, while another thread, the weft, is passed iteratively from side to side, i.e.\new{,} orthogonally to the warp. At this point it is interesting to remark that separations between threads in the warp presents a deterministic pattern that depends on how the threads have been arranged\RED{,} while in the weft the separation along the roll depends on how the threads have been tightened. This is the reason why the \RED{thread} counting, i.e.\new{,} the count of threads per cm, both in the vertical and horizontal axes, has been widely used as characteristic of the canvases. 
Besides, because of deformations of the fabric around nails in the stretcher persists in time\footnote{Prime is applied to the cloth after nailing, and this prime acts as a glue.}, deviations of threads respect to the horizontal and vertical axis are useful for the curator to study the painting and 
provide information on transformations of areas in the canvas, checking for integrity, and better understanding how the painter's workshop was organized in the productions of series. 
 
\begin{figure}[!tb]
\centering
\begin{tabular}{cccc}
 \includegraphics[width=3.1cm]{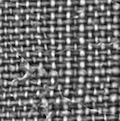}& 
  \includegraphics[width=3.1cm]{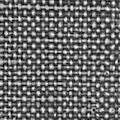}\\
(a) & (b) 
\end{tabular}
\caption{Scenarios where the FT fails, fabrics in canvases by (a) Rubens with different thread widths and (b) Velázquez where vertical threads are seen as \new{widenings} of horizontal ones at the crossing points.} \label{fig:Problems}
\end{figure}

The FT has been successfully applied to this problem of plain wave canvases analysis \cite{Johnson2013, Simois18}, exhibiting quite a robust performance. 
However, we found that the FT fails in two common scenarios. First, whenever threads are of different widths, for the pattern is not uniform, the FT detects several maxima. As a result, in nearby areas of the canvas the FT provides quite different threads densities, \new{since} it changes from one maximum to another as we process locations one next to the other. In Fig. \ref{fig:Problems}.(a) we include a \new{sample} of the X-ray of the canvas Adan and Eva from Rubens \cite{P001692}, where threads of several different widths are observed.  On the other hand, in some fabrics and in one direction, usually the warp, the threads are quite tight and the orthogonal ones are just perceived as widenings at the crossing points. In this scenario the FT is unable to provide any estimation of the thread density.  In Fig. \ref{fig:Problems}.(b) we bring a \new{sample} of the X-ray plate of the Prince Baltasar Carlos on Horseback by Velázquez (P001180 in MNP inventory) \cite{P001180}, horizontal threads are easily observed while the vertical ones can be only identified by focusing on the widening of the horizontal threads. 

\subsection{Pipeline and Contributions}
In the following, after describing related work in Sec. \ref{sec:RW}, we face the analysis of thread densities and angles in plain weaves by means of segmentation. As a first solution, we tried to segment vertical and horizontal threads independently. However, as fabrics \new{have severe rotations at some points,} the model was unable to distinguish a vertical (or horizontal) thread from a severely rotated horizontal (vertical) one. As a result, the learning did not converge and we decided to segment \cite{Shervin2022} crossing points instead, with just one model. Accordingly, the pipeline was designed as follows.

In the dataset generation, described in Sec. \ref{sec:data}, we used paintings from several authors in the 17th-19th centuries. From these paintings\RED{,} random \new{samples} were obtained and labeled. We annotated both the vertical and horizontal threads in every \new{sample}. The labeled samples were divided into training, validation and test subsets. 
We investigated several models, see Sec. \ref{sec:DL}\new{,} and adopted the U-Net with inception \cite{Inception14}. The Dice loss function was chosen, exhibiting good performance.  At this point, it is interesting to note that we do not have densities nor angle estimations\RED{,} but the locations of the crossing points. Hence, an algorithm to estimate these values from the crossing points was developed. This approach is described in Sec. \ref{sec:SC}. The results of the training of the proposed model and \new{its validation} are included in Sec. \ref{sec:train}. In production, we process a whole canvas to provide the densities and angle deviations at every point. Note that we are not using the trained model as transfer learning to analyze a new canvas, as this would involve a labeling process for every new painting \cite{Aradillas21}. We do not assume any annotation for a new canvas analysis \cite{Maaten15}. Sec. \ref{sec:exp} is devoted to \new{illustrating} the good performance of the developed approach compared to the method in the frequency domain. A \new{couple of pairs of canvases by Velázquez and Murillo are analyzed. We \RED{also include} the study of an X-ray plate of a painting by Ribera.}
We end with conclusions. 


%

The main contributions of the paper are as follows:
\begin{enumerate}
\itemsep=0pt
\item A description of problems of the FT applied to the thread counting problem.
\item A method to crop and label X-rays of paintings, augmenting the resulting dataset.
\item Four DL models to detect crossing points of threads in plain weave, working in the spatial domain. These DL models follow the U-Net architecture and have been programmed using Keras-\new{TensorFlow}. \new{We report that the segmentation of vertical, or horizontal, threads yields poor results.}
\item An approach to compute threads densities and angle estimations from crossing points.
\item A whole thread counting algorithm in the spatial domain with no need of labeling part of the to be processed canvas.
\item Analysis of the errors in the estimation of thread densities for labeled data of X-rays of different qualities and fabrics of different densities.
\item Application \new{to a pair of case studies with canvases by Velázquez and Murillo}, to conclude the fabrics used coming from the same roll, very much outperforming the FT approach.

\end{enumerate}  


\new{
\section{Related Work}\label{sec:RW}
Image processing has been applied to the study of priceless paintings \cite{Barni05,Cornelis17,Deligiannis2017,Johnson08}. 
Recently, algorithms in the machine learning (ML) field based on artificial neural networks have been applied. In 
\cite{Rucoba22} crack detection is solved by applying K-SVD and in \cite{Sizyakin20} convolutional neural networks (CNN) were used. CNN were also applied to automatic classification of paintings \cite{Roberto2020}. In \cite{Pu2020} auto-encoders (AE) were used for image separation. In \cite{Zou21} deep learning (DL) was applied for virtual restoration of \new{colored} paintings.  Authentication and forgery detection has \RED{also been} the focus of these techniques \cite{Polatkan09,Nemade2017}. Segmentation \RED{approaches based on U-Net \cite{Unet15} and AE \cite{AE86,Goodfellow2016} were} applied to image restoration by inpainting. 

Within the field of image processing and machine learning (ML) the study of the fabric has received particular attention. In \cite{Escofet2001} the theoretical frequency analysis of fabrics was introduced, being updated in \cite{Simois18}. Later, \new{this} theoretical background was translated into studies of canvases \cite{Johnson2010, Johnson2013} by means of the Fourier transform (FT). Another form of transform was also tried \cite{Yang2015} and the power spectral density was introduced as a feature of the canvas in \cite{Simois18}. These \new{transform-based} tools are unsupervised, as no previous labeling of the \new{threads} in the images or their densities was needed. These FT approaches are robust and useful to process many masterpieces, but they fail in some scenarios, as discussed earlier. 
In \cite{Maaten15} the authors presented a Bayesian tool to predict the positions of the crossing points in plain weave. This is, up to our knowledge, the only ML tool applied to the \new{thread} counting problem. However, a prior labeling stage was needed for the canvas at hand, where the curator needs to mark the crossing points for a given area of the to be processed canvas. In this manuscript we present a novel ML tool for plain weave thread counting to solve some drawbacks of the FT approaches, with no need of pre-labeling. 
}

\section{Data Preparation}\label{sec:data}

\subsection{Dataset}
We first selected 36 paintings from the Museo Nacional del Prado (MNP), of more than 13 different painters. Canvases from Rubens, Velázquez, Lorena, Swanevelt, Dughet, Poussin, Both, Lemaire, or Ribera were included in the dataset. These paintings were selected to be representative of several densities in the range 6 to 23 threads per cm (thr/cm), different resolutions of the image and noise conditions. This encompasses the usual densities found in canvases. For example, in the French paintings analyzed for the 17th-20th centuries and 44 painters the densities of the threads were in the range 6-23.3 thr/cm. For the 17th and 18th \new{centuries,} the range reduces to 8.6-20 thr/cm \cite{Vanderlip80}.
 
\begin{figure}[!tb]
\centering
\begin{tabular}{cccc}
 \includegraphics[width=3.6cm]{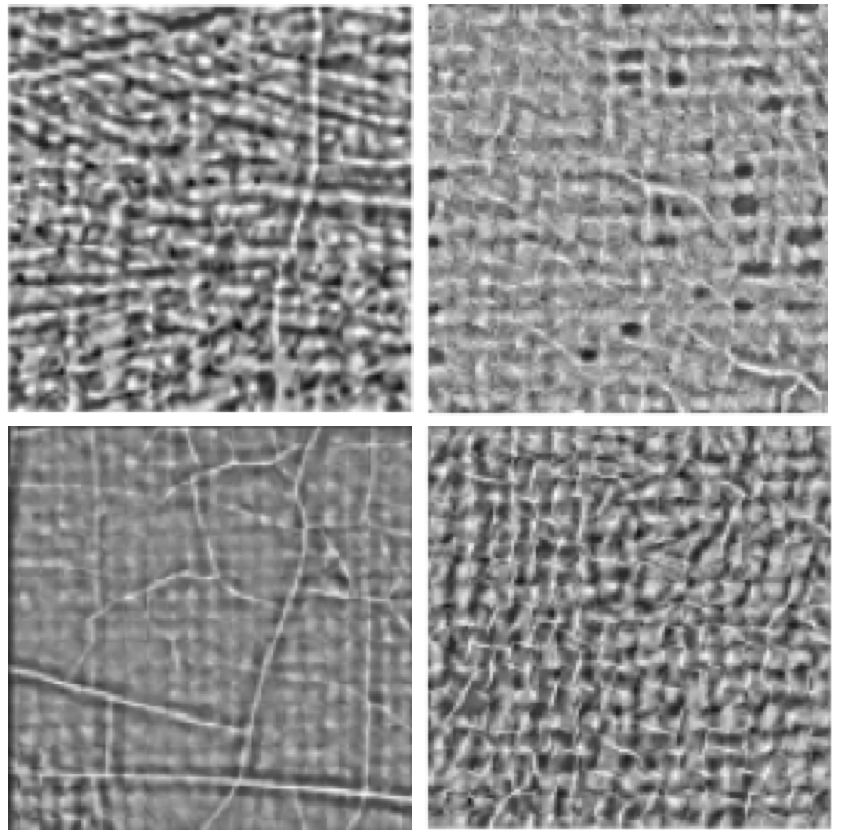}& 
  \includegraphics[width=3.6cm]{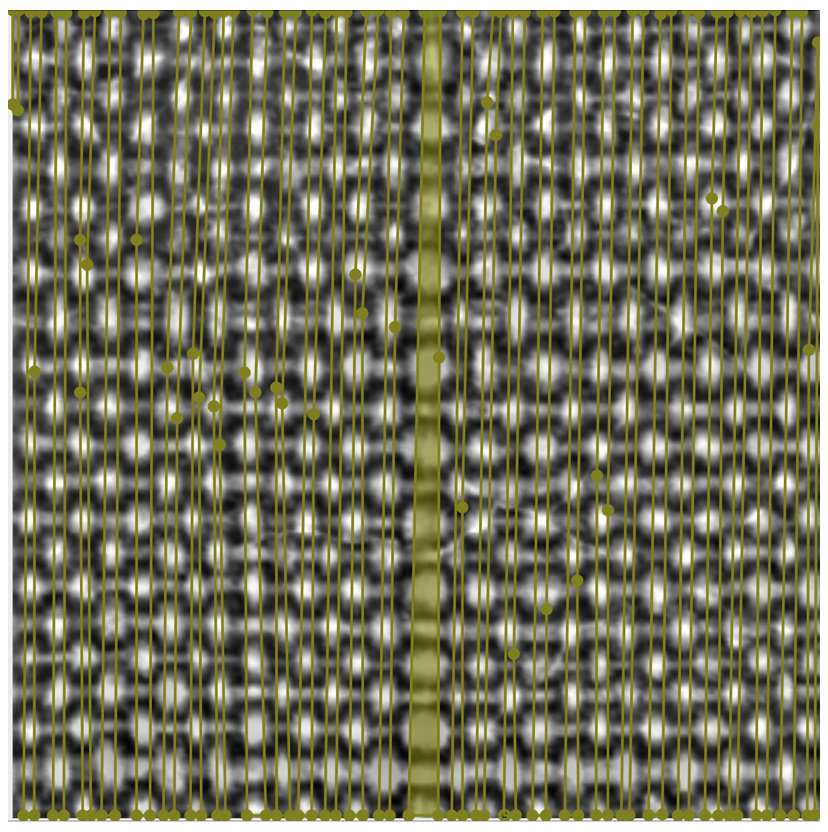}\\
(a) & (b) 
\end{tabular}
\caption{Examples of (a) four \new{samples where} the fabric is hardly observed and (b) a labeled crop with annotations for vertical threads, one of the labeled threads is highligthed.} \label{fig:crops}
\end{figure}

The images of the X-ray plates were obtained after digitalization\RED{,} with resolutions ranging from 80 to 200 pixels per cm (ppc). The images were enhanced with algorithms based on their local mean and variance \cite{Lee80} then scaled, \new{see Appendix \ref{sec:Prep} for a description \RED{of} these steps}. For every plate we \new{cropped 40 random areas of $1.5 \times 1.5$ cm} and resampled them, if needed, to 200 ppc. For every canvas, approximately \new{7 samples out of the 40} were labeled. This number was reduced or increased if its densities and quality were already represented or not by other similar fabrics in the dataset. Overall, we labeled 239 \new{samples}. The labeling of these \new{samples} was performed not for the crossing points but for the vertical and horizontal threads independently, then a unique labeled image was obtained with the crossing points. This stage was specially elaborated as in many cases the fabric was hardly observed in the X-ray plate, see four different \new{samples} in Fig. \ref{fig:crops}.(a). For high densities, the labeling becomes challenging as the space between threads is quite narrow, if any. See Fig. \ref{fig:crops}.(b), where the annotations for one \new{sample} is included \new{and} one of the annotations is highlighted. \new{More} than one person participated in the labeling. Therefore\new{,} for the same canvas we may have slightly different widths for the annotations. 

\new{Then}, we divided the dataset into training, validation and test. Instead of randomly distributing the \new{samples} into these three groups, we took the labeled \new{samples} from 4 paintings for validation, 31 \new{samples} overall,  and the samples of 3 other paintings for test purposes, with 18 labeled \new{samples}.  Hence, out of the available paintings, 29 were used for training, 4 for validation and 3 for test. 
We enforced that not only different paintings were used in each group\RED{,} but that canvases with labeled fabrics known to come from the same roll were placed in the same subset. Paintings in the validation and test datasets were selected to be representative of different densities and qualities. At this point\RED{,} it is important to remark that these are not the final \new{samples} used. Next\new{,} we explain how to generate the \new{samples} fed to the model for the training, validation and test stages.


%

\subsection{Data Generation and Augmentation}

We need a large dataset to represent the different possible inputs. In our design\RED{,} we labeled \new{$1.5\times 1.5$ cm samples at 200 ppc resolution. However,} the model, as explained later, is fed with 1 $\times$ 1 cm \new{images} (200 $\times$ 200 pixels).  
The first step to generate the input datasets was to obtain \new{these} images from \new{every sample}. We performed this task by \new{first} taking the corners of the \new{sample} to get the first 4 \new{images}. Then four rotated versions of two \new{images} given by the (50:250, 50:250)  and (65:265, 65:265) pixels \new{of the sample} were added, with random angles in the ranges $[2,7]$, $[-2,-7]$, $[8,12]$, $[-8,-12]$. We also added two rotated versions of the \new{images} given by the (65:265, 65:265) pixels \new{of the sample}, with random angles in the ranges $[2,7]$, $[-2,-7]$.  This process yield $10$ samples for every \new{sample}.
The dataset was augmented \cite{DA2019} by repeating the whole process after a) left-right flipping, b) up-down flipping, c) rotating $90^\circ$, d) rotating $90^\circ$ plus left-right flipping and e) rotating $90^\circ$ plus up-down flipping the full \new{sample}. 
At the end of the whole process we had $60$ \new{images} of size $200 \times 200$ for every labeled $300\times300$ pixel \new{sample}. Since we labeled 239, we generated 14340 \new{images}. 

The resulting dataset was slightly skewed. Some of the labeled \new{samples} came from canvases with similar fabrics, hence increasing the number of samples of the same type. To prevent the model to overfit to them\RED{,} we generated twice the samples for the other paintings, including them in the dataset. By repeating the generation of samples for these fabrics we increased the number of \new{images} to $21540$.

\section{DL Model}\label{sec:DL}
\begin{figure*}[htp]
\centering
 \includegraphics[width=1\linewidth]{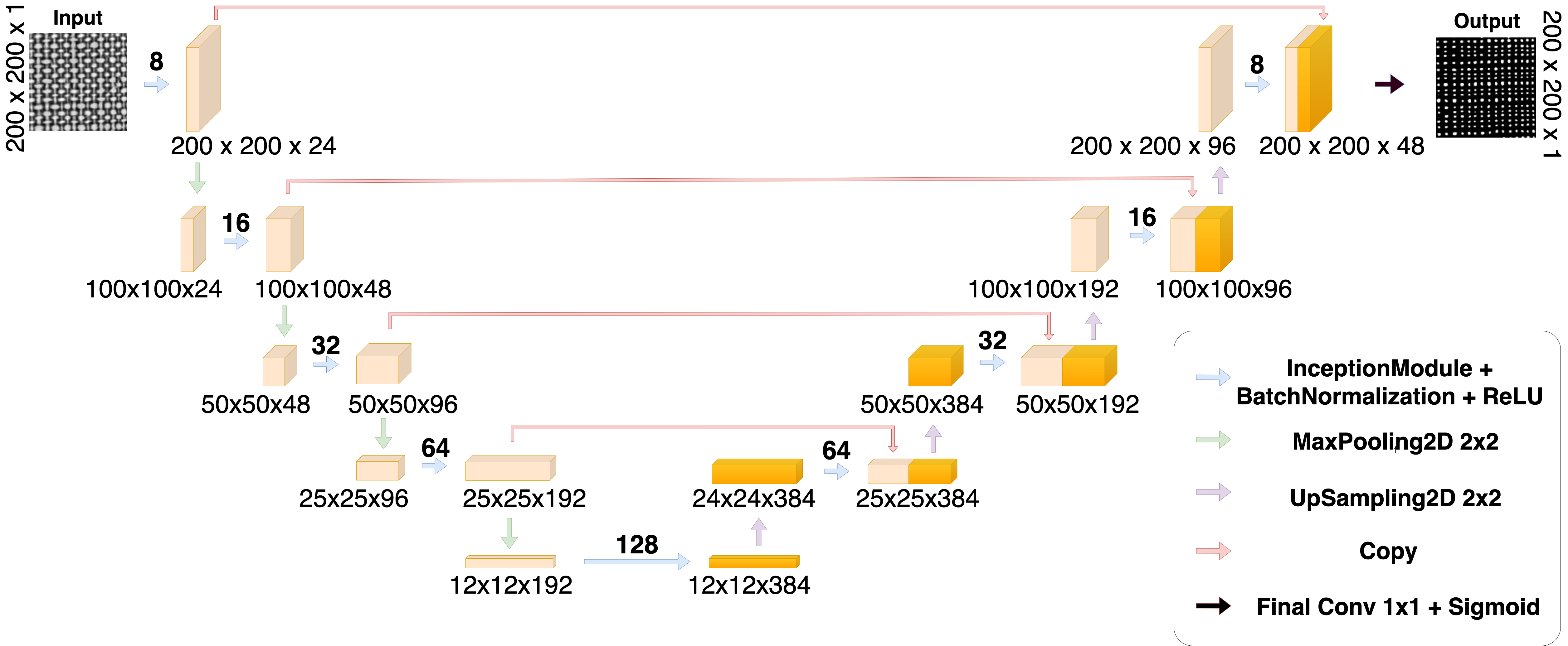}
\caption{The U-Net architecture used with the inception modules in blue horizontal arrows, with the number of filters used, $n_i$, indicated on top. Up and down steps involve halving reductions of sizes downwards, and \new{double} them upwards, with zero padding if needed. Skip \new{(copy)} connections \new{concatenate} the upcoming features with the features at the same level in the encoder. The size of all feature maps is indicated below them. At the output a $1\times1$ convolutional filter is used, with sigmoid activation function.} \label{fig:Unet}
\end{figure*}

We designed a DL model to detect and segment crossing points between vertical and horizontal threads. The U-Net architecture \cite{Unet15,Ali2022} was the starting point to build the model, see Fig. \ref{fig:Unet}, where a top-down path, usually denoted as encoder or contracting path, is followed by a set of down-top layers, the decoder or expansive path. Layers are composed of one main unit, i.e., \new{a module with 2D convolutions,} batch normalization and a ReLU activation function. \new{We developed several models based on this architecture. In the following\RED{,} we first describe the best model found, i.e., the one providing the best threads densities estimations for the validation subset. Then we outline other evaluated models.}

\subsection{Inception Module}\label{ssec:incDice}

\begin{figure}[htp]
\centering
 \includegraphics[width=10cm]{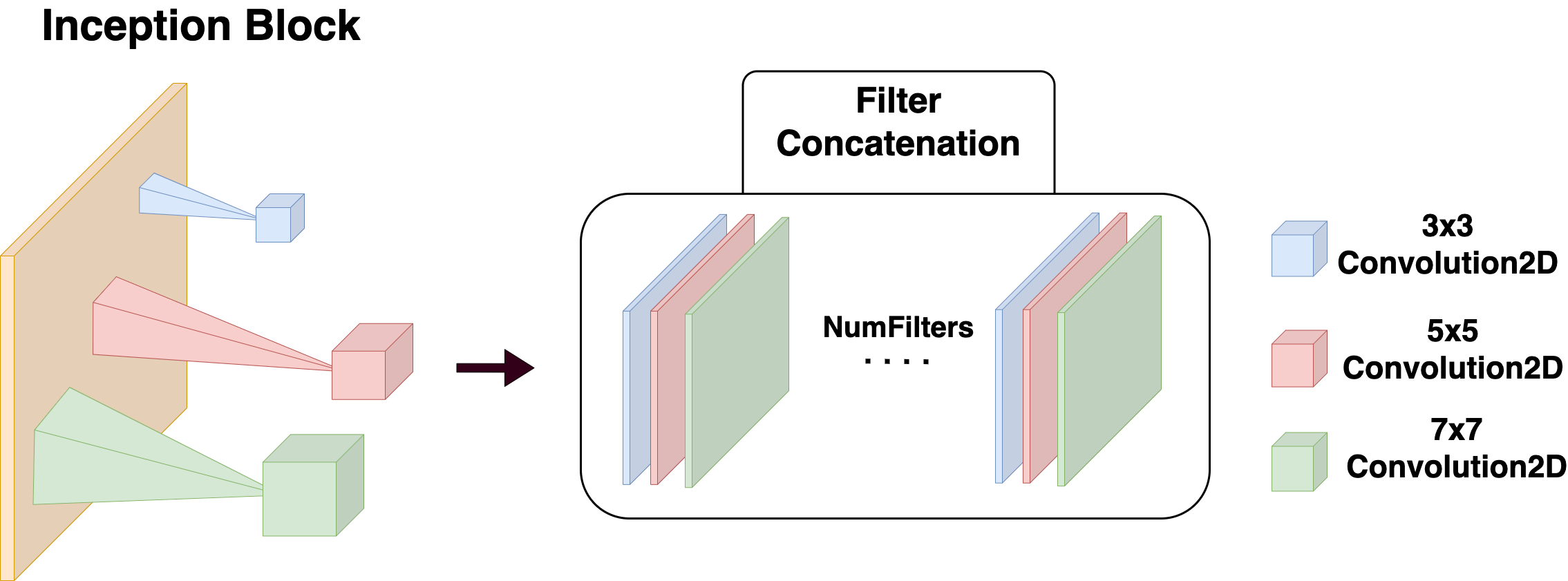}
\caption{Inception block: $3\times3$, $5\times5$ and $7\times7$ kernels are convolved with the input at the same depth, as many times as given by the number of filters parameter, $n_i$. The results of these convolutions are concatenated.} \label{fig:inception}
\end{figure}

We first tried a fixed size 2D convolutional kernel in every layer, while the size could change from \new{one layer to the next}.  However, the outcome of this model was improved by introducing inception. Our conjecture is that \new{the variety of threads densities of the canvases,} in the range 6 to 23 thr/cm, makes it hard for the model to locate the crossing points if one fixed-size kernel is used. On the contrary, by means of the inception paradigm, in the same layer we have convolutional kernels of several sizes. \new{We used $k\times k$ kernels with $k$=3, 5 and 7}, see Fig. \ref{fig:inception}. Note that similar ideas \new{were} adopted in \cite{Zhang2022,Ali2022,Yamanakkanavar2022}. The inception module, see horizontal (blue) arrows within the encoder and decoder in Fig. \ref{fig:Unet}, is a sequence of blocks where the first one is a convolution with $n_i\times3$ different kernels,  where $n_i$ denotes the number of filters per layer and $i$ the layer level from top to bottom, see Fig. \ref{fig:inception}. The results are concatenated at the output. Then a batch normalization is performed, followed by a ReLU activation function. We used $n_0=8$ filters for the first upper layer and scaled this number by $2$ as we went deeper in the encoder, $n_{i+1}=2n_{i}$. 
\new{All convolutions were performed with the `same' option, i.e., zero padding of size $\lfloor{k/2}\rfloor$ was introduced around every input image to get the same size at the output after convolution.}



\subsection{Model Description}\label{ssec:incDicemodel}


\new{The model used, in Fig. \ref{fig:Unet}, has an image of $200\times200\times 1$ at the input layer. Then, in the encoder stage, the input undergoes 5 layers down to get a tensor of dimensions $12\times12\times 384$. Each layer is composed of the following blocks:
\begin{enumerate}
\item Inception: three parallel 2D convolutional layers with square kernels of size 3, 5 and 7. Outputs of the inception are concatenated at the output, the number of features at the output of the inception is $3$ times the number on top of the horizontal arrows in the encoder, that indicates the number of inception kernels. 
\item Batch normalization.
\item ReLU activation function.
\item Max-pooling of stride 2 and kernel size 2, except for the last (5th) layer of the encoder.
\item Dropout of probability $p=0.1$.
\end{enumerate} 
After the encoder\RED{,} the resulting features go up through 4 layers in the decoder stage. Each layer in the decoder includes the following:
\begin{enumerate}
\item Upsampling of size $2\times 2$, except for the last layer in the decoder (9th layer in the full network).
\item Inception: three parallel 2D convolutional layers with square kernels of size 3, 5 and 7. Outputs of the inception are concatenated at the output, the number of features at the output of the inception is $3$ times the number on top of the horizontal arrows in the decoder, that indicates the number of inception kernels. 
\item Batch normalization.
\item ReLU activation function.
\item Copy and concatenate the tensor at the output of the layer at the same level (same size) at the encoder, see long horizontal arrows joining the two vertical paths of the ``U'' shape in Fig. \ref{fig:Unet}.
\item Dropout of probability $p=0.1$.
\end{enumerate}
Finally, at the output layer we have a 2D convolutional layer of kernel size 1 and a sigmoid as activation function.
The outputs of the first 5 layers in the encoder have sizes \RED{$100\times100\times48$}, $50\times50\times96$, $25\times25\times192$, $12\times12\times192$ and $12\times12\times384$, that are, in reverse order, the sizes of the inputs to the layers at the same level in the decoder and the output layer. The full model has over 6 million parameters.
}

\subsection{Loss Function}\label{ssec:incDiceloss}

\begin{figure}[htb]
\centering
\begin{tabular}{cccc}
 \includegraphics[width=3.0cm]{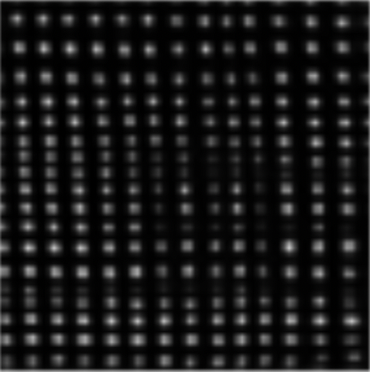}& 
  \includegraphics[width=3.0cm]{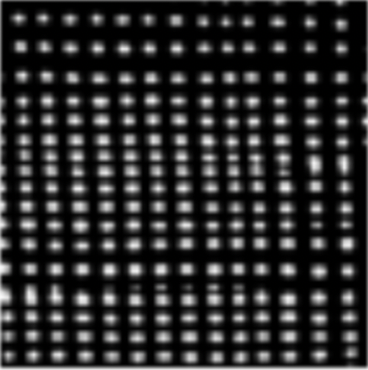}\\
(a) & (b) 
\end{tabular}
\caption{Segmented image obtained with (a) binary cross-entropy as loss function and (b) Dice loss.} \label{fig:Dice}
\end{figure}

The binary cross-entropy was first used as loss function. However, when analyzing an image the output values \new{of the model} were far from being binarized and a thresholding was needed, where the Otsu \cite{Otsu79} method was applied. To \new{provide an output with more extreme values} we better used the Dice loss, then in the analysis stage the image was binarized with \new{just} a $0.5$ threshold.  
\new{
Given the labeled output $\ma{I}$, a matrix of bits, and the estimated one $\hat{\ma{I}}$, the Dice loss is given by:
\begin{equation}\label{eq:Dice}
  L_D = 1 - \frac{2\sum_{i,j}   \ma{I}[i,j] \hat{\ma{I}}[i,j]} {\sum_{i,j}   \ma{I}[i,j] +\sum_{i,j}   \hat{\ma{I}}[i,j] } 
\end{equation}
}
In Fig. \ref{fig:Dice} we included the output for a segmentation 
\new{after training with} 
binary cross-entropy and Dice loss. Note that the resulting output with Dice loss is a quasi-binary image, as needed later to estimate densities and angles of threads.  We also observed better results in accuracy, used as validation loss, see Sec. \ref{sec:train}.
%
\new{
Therefore, in the learning stage of the model presented above, later used in the analysis of some case studies, we used the Dice loss in (\ref{eq:Dice}) as error to train the parameters of the networks and the accuracy as validation loss.
}

\subsection{Other Models}
We described above the model that provided the best results in validation and the one used in the studies later included. Other three models were also \new{proposed.} The four of them share the number of layers, the \new{max-pooling} at every layer in the encoder, the sigmoid as activation layer in the output of the model and the ReLU as activation function in the other layers. 
The accuracy was used in all cases to validate the model. The main features of these other models \new{are described in this subsection}.

\subsubsection{U-Net with Otsu}\label{ssec:UnetTh}
The U-Net model with layers of one kernel size was the first tried model. At the output\RED{,} a threshold computed using the Otsu approach \cite{Otsu79} was applied prior to performing the thread densities estimations. In the encoder\RED{,} the layer unit was a double repetition of convolutional plus batch normalization and ReLU blocks. All kernels size where $k=3$ except for the first layer, set to $k=7$. \new{The} initial number of filter was $n=14$ and this number was doubled in the next layer, $n_{i+1}=2n_{i}$. The dropout was set to $p=0.25$ and the learning rate to $10^{-4}$. These values were the ones exhibiting the best results. In the decoder\RED{,} the \new{transposed convolution} was applied to upsample the intermediate results. The loss function used was the binary cross-entropy. 

\subsubsection{U-Net with Dice}\label{ssec:UnetDice}
This model was equivalent to the previous one but for the loss function, where the Dice in (\ref{eq:Dice}) was used instead. As the result was already quite binarized a $0.5$ threshold was used at the output, to estimate the densities. The learning rate was increased to $10^{-3}$.

\subsubsection{Original Inception}\label{ssec:OrigUnetDice} 
Another model used was a closer version of the inception model in \cite{Inception14}. As inception module we included four parallel convolutional layers:
\begin{enumerate}
\item A $k=1$ convolution with batch normalization and ReLU.
\item A $k=1$ convolution with batch normalization and ReLU followed by a $k=3$ convolution with batch normalization and ReLU.
\item  A $k=1$ convolution with batch normalization and ReLU followed by a $k=3$ convolution, another $k=3$ convolution, batch normalization and ReLU.
\item A \new{max-pooling} of size $3\times3$ but with stride one, followed by a $k=1$ convolution with batch normalization and ReLU.
\end{enumerate}
The output of these four sequences of blocks were concatenated at the output. The learning rate was set to $10^{-3}$.

\subsection{SW and HW Details}

The models described above were programmed using Python 3.9 and Keras-Tensorflow 2.5.0. The input grayscale image size was $200\times200\times1$ and the default batch size was 32. We used the \RED{Adam} optimizer \cite{Kingma15} with a learning rate \RED{value} of $10^{-3}$. An early stopping was also included with a latency of 5. \new{A Tesla P100 with 16GB memory was used along with a Intel Xeon E5-2630 v4 with 20 cores CPU.}

\section{Density and Angle Estimations}\label{sec:SC}

\new{At the output of our model we have an image with high values at the crossing points areas, see Fig. \ref{fig:Dice}.(b) for an example, where a $1$ cm side image is included. After binarization we locate these areas by computing their centroids. Next, we propose to estimate the densities of the vertical and horizontal threads as described in Algorithm \ref{alg:SC},
where $\overline{\vc{h}}$ denotes the mean value of vector $\vc{h}$ and we exploited the fact that inputs images have a resolution of $200$ pixeles per cm. In this work we used $m=9$, $\alpha=25^\circ$ and $q=10$ or $q=0$. 
}
\new{
For every crossing point found, the method searches for the $m$ nearby segmented crossing points to estimate distances and angles. In Fig. \ref{fig:spatialCountProcess} we include an example where for a crossing point, circled in red, the $m=9$ nearest ones are found.
Within these $m$ neighbors, the nearest ones above ($+90^\circ$), below ($-90^\circ$), to the right ($0^\circ$) and to the left ($180^\circ$) are found, if any. See Fig. \ref{fig:spatialCountProcess} where some of these points have been marked with arrows. The averaged distances to the neighbors for all the crossing points are used to estimate the thread densities. 
Before averaging, removing values below and above a percentile, given by parameter $q$, might eliminate possible outliers. Angle orientations of the vertical and horizontal threads can also be estimated by checking the angles of the vectors pointing to the nearest neighbors.  
%

\begin{figure}[htp]
\centering
 \includegraphics[width=3.5cm]{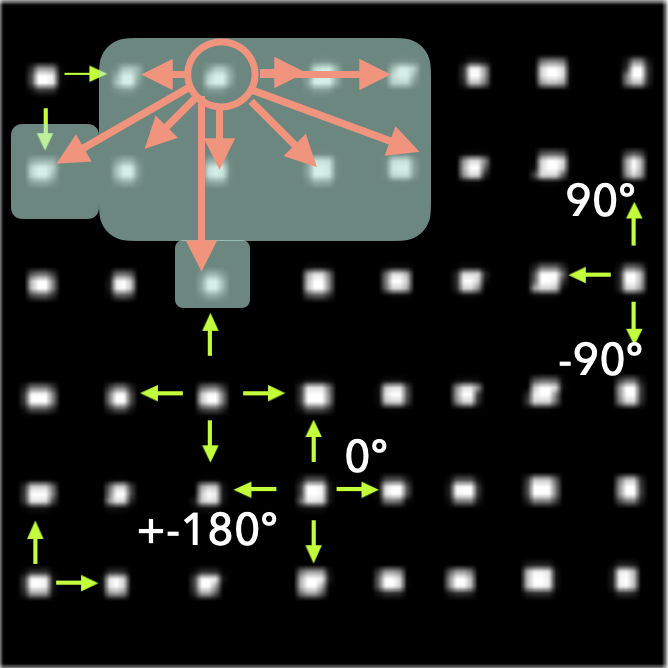}
\caption{\new{Sketch of the search for $m=9$ neighbors for a given crossing point (circled). The arrows to the neighbors indicate the vectors used to compute the angle deviations and the distances.}} \label{fig:spatialCountProcess}
\end{figure}
}

\begin{algorithm}
\caption{Spatial Counting (SC)}\label{alg:SC}
\textbf{Input:} the number of segmented crossing points, $M$, and the centroids of the crossing points.
\begin{algorithmic}
\For {k = 1 to $M$} estimate the $m$ nearest centroids to the $k$th one and store the distances to them, in pixels, as the $k$th row of the $M\times m$ matrix $\ma{D}$. 
\EndFor
\For {k = 1 to $M$} estimate the angles of the vectors joining the $k$th centroid to the $m$ nearest ones and store the result as the $k$th row of the $M\times m$ matrix $\ma{A}$. 
\EndFor
\State Initialize  $\vc{h}$ and $\vc{v}$ as empty vectors.
\For {k = 1 to $M$}
\State Search for the entries of the $k$th row of $\ma{A}$ with values in the range $[90-\alpha,90+\alpha]$ and, if any, store in $\vc{v}$ the lowest distance found within the corresponding entries of the $k$th row of  $\ma{D}$. 
\State Search for the entries of the $k$th row of $\ma{A}$ with values in the range $[-90-\alpha,-90+\alpha]$ and, if any, store in $\vc{v}$ the lowest distance found within the corresponding entries of the $k$th row of  $\ma{D}$. 
\State Search for the entries of the $k$th row of $\ma{A}$ with values in the range $[-\alpha,+\alpha]$ and, if any, store in $\vc{h}$  the lowest distance found within the corresponding entries of the $k$th row of  $\ma{D}$. 
\State Search for the entries of the $k$th row of $\ma{A}$ with values in the range $[180-\alpha,180+\alpha]$ and, if any, store in $\vc{h}$ the lowest distance found within the corresponding entries of the $k$th row of  $\ma{D}$.  
\EndFor
\State Remove from $\vc{h}$ and $\vc{v}$ values below the $q$th percentile and above the ($100-q$)th percentile. 
\State Compute the horizontal, $h$, and vertical, $v$, thread densities, in threads/cm, as 
\begin{align}
h &= 200 / \overline{\vc{h}} \\
v  &= 200 / \overline{\vc{v}} 
\end{align}
Likewise, the average tilt of the vertical and horizontal threads can be calculated by means of the average of all the angles of the corresponding crossing points in $\vc{v}$ and $\vc{h}$, respectively.
\end{algorithmic}

\end{algorithm}

Another approach to estimate the densities could be applying frequency analysis, i.e., the FT, to the result of the segmentation. We will refer to this method as FA.

\section{Training and Testing}\label{sec:train}

\subsection{Training Models}\label{ssec:modelsel}
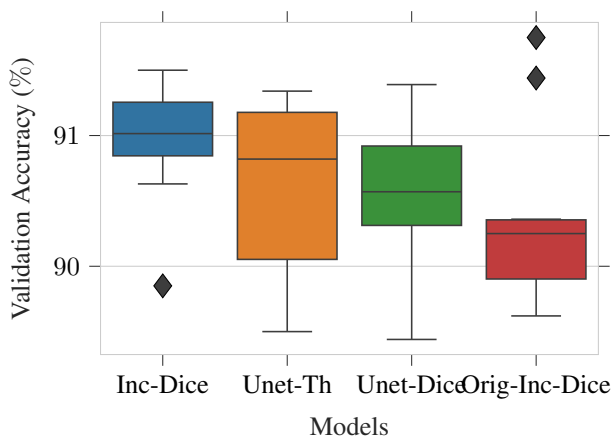
\begin{figure}[htp]
\centering
\begin{tikzpicture}

\definecolor{brown1926061}{RGB}{192,60,61}
\definecolor{darkslategray38}{RGB}{38,38,38}
\definecolor{darkslategray61}{RGB}{61,61,61}
\definecolor{lightgray204}{RGB}{204,204,204}
\definecolor{peru22412844}{RGB}{224,128,44}
\definecolor{seagreen5814558}{RGB}{58,145,58}
\definecolor{steelblue49115161}{RGB}{49,115,161}

\begin{axis}[
width=8.2cm,
height = 6cm,
axis line style={lightgray204},
tick align=outside,
x grid style={lightgray204},
xlabel=\textcolor{darkslategray38}{Models},
xmajorticks,
xmin=-0.5, xmax=3.5,
xtick style={color=darkslategray38},
xtick={0,1,2,3},
xticklabels={Inc-Dice,Unet-Th,Unet-Dice,Orig-Inc-Dice},
y grid style={lightgray204},
ylabel=\textcolor{darkslategray38}{Validation Accuracy ($\%$)},
ymajorgrids,
ymin=89.3245, ymax=91.8655,
ytick style={color=darkslategray38}
]
\path [draw=darkslategray61, fill=steelblue49115161, semithick]
(axis cs:-0.4,90.845)
--(axis cs:0.4,90.845)
--(axis cs:0.4,91.255)
--(axis cs:-0.4,91.255)
--(axis cs:-0.4,90.845)
--cycle;
\path [draw=darkslategray61, fill=peru22412844, semithick]
(axis cs:0.6,90.0525)
--(axis cs:1.4,90.0525)
--(axis cs:1.4,91.1775)
--(axis cs:0.6,91.1775)
--(axis cs:0.6,90.0525)
--cycle;
\path [draw=darkslategray61, fill=seagreen5814558, semithick]
(axis cs:1.6,90.3125)
--(axis cs:2.4,90.3125)
--(axis cs:2.4,90.92)
--(axis cs:1.6,90.92)
--(axis cs:1.6,90.3125)
--cycle;
\path [draw=darkslategray61, fill=brown1926061, semithick]
(axis cs:2.6,89.9025)
--(axis cs:3.4,89.9025)
--(axis cs:3.4,90.355)
--(axis cs:2.6,90.355)
--(axis cs:2.6,89.9025)
--cycle;
\addplot [semithick, darkslategray61]
table {%
0 90.845
0 90.63
};
\addplot [semithick, darkslategray61]
table {%
0 91.255
0 91.5
};
\addplot [semithick, darkslategray61]
table {%
-0.2 90.63
0.2 90.63
};
\addplot [semithick, darkslategray61]
table {%
-0.2 91.5
0.2 91.5
};
\addplot [black, mark=diamond*, mark size=4.5, mark options={solid,fill=darkslategray61}, only marks]
table {%
0 89.85
};
\addplot [semithick, darkslategray61]
table {%
1 90.0525
1 89.5
};
\addplot [semithick, darkslategray61]
table {%
1 91.1775
1 91.34
};
\addplot [semithick, darkslategray61]
table {%
0.8 89.5
1.2 89.5
};
\addplot [semithick, darkslategray61]
table {%
0.8 91.34
1.2 91.34
};
\addplot [semithick, darkslategray61]
table {%
2 90.3125
2 89.44
};
\addplot [semithick, darkslategray61]
table {%
2 90.92
2 91.39
};
\addplot [semithick, darkslategray61]
table {%
1.8 89.44
2.2 89.44
};
\addplot [semithick, darkslategray61]
table {%
1.8 91.39
2.2 91.39
};
\addplot [semithick, darkslategray61]
table {%
3 89.9025
3 89.62
};
\addplot [semithick, darkslategray61]
table {%
3 90.355
3 90.36
};
\addplot [semithick, darkslategray61]
table {%
2.8 89.62
3.2 89.62
};
\addplot [semithick, darkslategray61]
table {%
2.8 90.36
3.2 90.36
};
\addplot [black, mark=diamond*, mark size=4.5, mark options={solid,fill=darkslategray61}, only marks]
table {%
3 91.75
3 91.44
};
\addplot [semithick, darkslategray61]
table {%
-0.4 91.015
0.4 91.015
};
\addplot [semithick, darkslategray61]
table {%
0.6 90.82
1.4 90.82
};
\addplot [semithick, darkslategray61]
table {%
1.6 90.57
2.4 90.57
};
\addplot [semithick, darkslategray61]
table {%
2.6 90.25
3.4 90.25
};
\end{axis}

\end{tikzpicture}
\caption{Box diagram of accuracy results for the four models trained: Inc-Dice, Unet-Th, Unet-Dice and Orig-Inc-Dice.} \label{fig:ValAcc}
\end{figure}

\new{We trained the proposed models ten times with different random initialization of the parameters}, i.e., the kernels, for A) the inception with Dice loss function (Inc-Dice) in Subsec. \ref{ssec:incDice}-\label{ssec:incDiceloss}, B) the U-net with Otsu threshold (Unet-Th) in Subsec. \ref{ssec:UnetTh}, C) the U-Net with Dice (Unet-Dice) in Subsec. \ref{ssec:UnetDice} and D) the original inception with Dice in Subsec. \ref{ssec:OrigUnetDice} (Orig-Inc-Dice).  
%
%
%
In Fig. \ref{fig:ValAcc} we include a box diagram with the accuracy results \new{of the segmentation} for the validation set after training\new{, with $q=10$}. The diamonds indicate outliers. It can be observed that the inception with Dice loss provides the best values, in mean and Q1 and Q3 quartiles. The \new{Unet-Th} does a good job, although with a higher dispersion. This dispersion is reduced with the \new{Unet-Dice}. The original inception exhibits a much worse result in most of the trainings, being the most computationally demanding, in view of the description in Subsec. \ref{ssec:OrigUnetDice}. 

\begin{figure}[htp]
\centering
\begin{tikzpicture}

\definecolor{brown1926061}{RGB}{192,60,61}
\definecolor{darkslategray38}{RGB}{38,38,38}
\definecolor{darkslategray61}{RGB}{61,61,61}
\definecolor{lightgray204}{RGB}{204,204,204}
\definecolor{peru22412844}{RGB}{224,128,44}
\definecolor{seagreen5814558}{RGB}{58,145,58}
\definecolor{steelblue49115161}{RGB}{49,115,161}

\begin{axis}[
width=8.2cm,
height = 6cm,
axis line style={lightgray204},
tick align=outside,
x grid style={lightgray204},
xlabel=\textcolor{darkslategray38}{Models},
xmajorticks,
xmin=-0.5, xmax=3.5,
xtick style={color=darkslategray38},
xtick={0,1,2,3},
xticklabels={Inc-Dice,Unet-Th,Unet-Dice,Orig-Inc-Dice},
y grid style={lightgray204},
ylabel=\textcolor{darkslategray38}{Validation Percent Error (\%)},
ymajorgrids,
ymin=0.926, ymax=5.194,
ytick style={color=darkslategray38}
]
\path [draw=darkslategray61, fill=steelblue49115161, semithick]
(axis cs:-0.4,1.21)
--(axis cs:0.4,1.21)
--(axis cs:0.4,1.3225)
--(axis cs:-0.4,1.3225)
--(axis cs:-0.4,1.21)
--cycle;
\path [draw=darkslategray61, fill=peru22412844, semithick]
(axis cs:0.6,1.3325)
--(axis cs:1.4,1.3325)
--(axis cs:1.4,2.465)
--(axis cs:0.6,2.465)
--(axis cs:0.6,1.3325)
--cycle;
\path [draw=darkslategray61, fill=seagreen5814558, semithick]
(axis cs:1.6,1.355)
--(axis cs:2.4,1.355)
--(axis cs:2.4,1.5575)
--(axis cs:1.6,1.5575)
--(axis cs:1.6,1.355)
--cycle;
\path [draw=darkslategray61, fill=brown1926061, semithick]
(axis cs:2.6,1.4175)
--(axis cs:3.4,1.4175)
--(axis cs:3.4,1.5475)
--(axis cs:2.6,1.5475)
--(axis cs:2.6,1.4175)
--cycle;
\addplot [semithick, darkslategray61]
table {%
0 1.21
0 1.12
};
\addplot [semithick, darkslategray61]
table {%
0 1.3225
0 1.33
};
\addplot [semithick, darkslategray61]
table {%
-0.2 1.12
0.2 1.12
};
\addplot [semithick, darkslategray61]
table {%
-0.2 1.33
0.2 1.33
};
\addplot [black, mark=diamond*, mark size=4.5, mark options={solid,fill=darkslategray61}, only marks]
table {%
0 1.65
};
\addplot [semithick, darkslategray61]
table {%
1 1.3325
1 1.3
};
\addplot [semithick, darkslategray61]
table {%
1 2.465
1 3.86
};
\addplot [semithick, darkslategray61]
table {%
0.8 1.3
1.2 1.3
};
\addplot [semithick, darkslategray61]
table {%
0.8 3.86
1.2 3.86
};
\addplot [black, mark=diamond*, mark size=4.5, mark options={solid,fill=darkslategray61}, only marks]
table {%
1 5
};
\addplot [semithick, darkslategray61]
table {%
2 1.355
2 1.24
};
\addplot [semithick, darkslategray61]
table {%
2 1.5575
2 1.7
};
\addplot [semithick, darkslategray61]
table {%
1.8 1.24
2.2 1.24
};
\addplot [semithick, darkslategray61]
table {%
1.8 1.7
2.2 1.7
};
\addplot [black, mark=diamond*, mark size=4.5, mark options={solid,fill=darkslategray61}, only marks]
table {%
2 1.96
};
\addplot [semithick, darkslategray61]
table {%
3 1.4175
3 1.37
};
\addplot [semithick, darkslategray61]
table {%
3 1.5475
3 1.66
};
\addplot [semithick, darkslategray61]
table {%
2.8 1.37
3.2 1.37
};
\addplot [semithick, darkslategray61]
table {%
2.8 1.66
3.2 1.66
};
\addplot [semithick, darkslategray61]
table {%
-0.4 1.285
0.4 1.285
};
\addplot [semithick, darkslategray61]
table {%
0.6 1.395
1.4 1.395
};
\addplot [semithick, darkslategray61]
table {%
1.6 1.41
2.4 1.41
};
\addplot [semithick, darkslategray61]
table {%
2.6 1.505
3.4 1.505
};
\end{axis}

\end{tikzpicture}
\caption{Box diagram of normalized absolute errors in the density estimations for the four models trained: Inc-Dice, Unet-Th, Unet-Dice and Orig-Inc-Dice. } \label{fig:ValErr}
\end{figure}
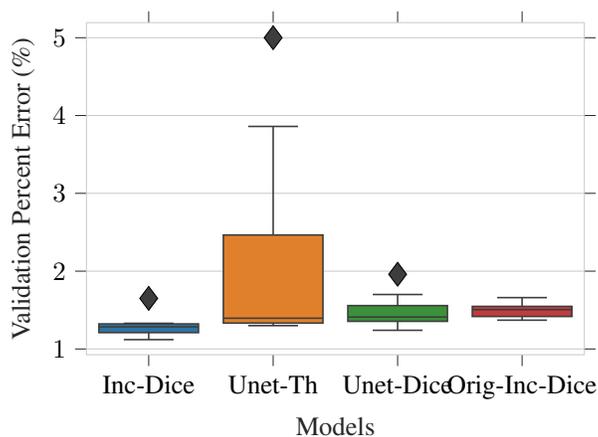

In Fig. \ref{fig:ValErr} the box diagram is represented for the error in the counting using the SC approach with $q=0$ for the trained models and validation set. The values for the horizontal and vertical densities were compared to the ones after applying the same SC approach to the ground truth (annotated samples with crossing-points).  We measured the mean of the absolute error normalized by the true value, to properly highlight errors in high densities. 
As expected, the median values increased in the same order as the validation decreased for the Inc-Dice, Unet-Th, Unet-Dice and Orig-Inc-Dice approaches. The training providing the lowest value was in the Inc-Dice set, \new{with} a $1.12\%$ error. In the following we will \new{use the Inc-Dice with these weights and SC, hereafter denoted by DLSC, to evaluate the performance in the test set and to analyze the case studies proposed.}

\new{The number of epochs varies from one training to the other. For the Inc-Dice the averaged number of epochs run was $8.7+5\approx 14$ where $5$ stands for the patience used in the early stopping. Since the training of the Inc-Dice lasted 180 s per epoch, each training of this model took approximately $42$ minutes. }


\subsection{Test Results}\label{ssec:test}
\begin{figure}[htp]
\centering
\begin{tabular}{ccc}
 \includegraphics[width=2.3cm]{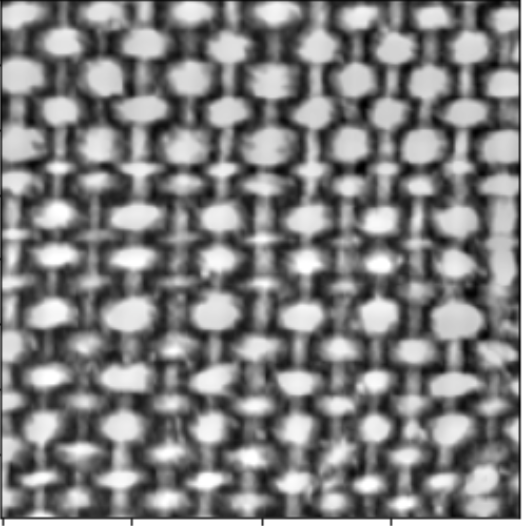}& \includegraphics[width=2.3cm]{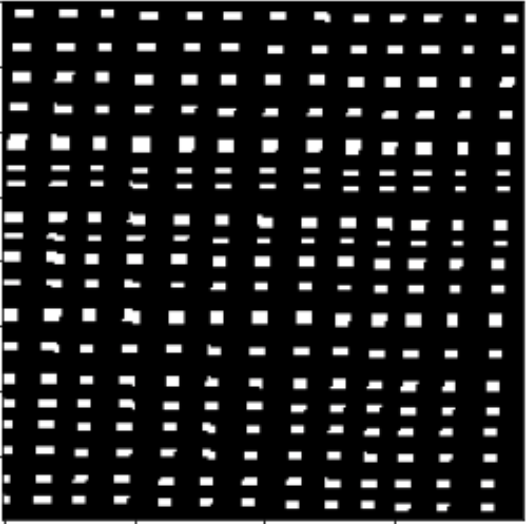}& \includegraphics[width=2.3cm]{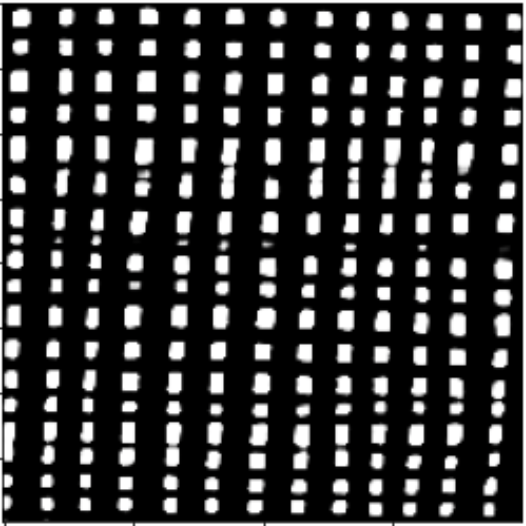} \\
 (a)&(b)&(c)\\
 \includegraphics[width=2.3cm]{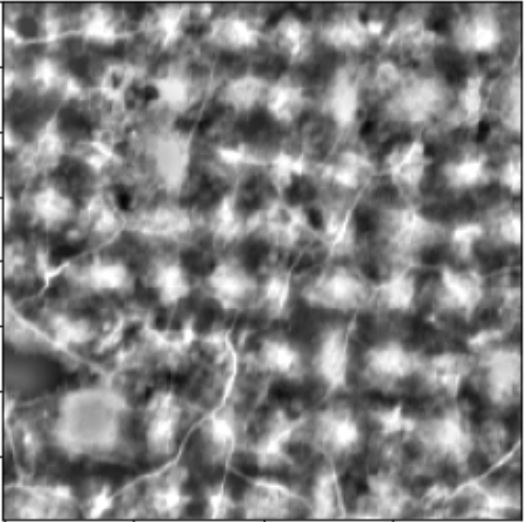}& \includegraphics[width=2.3cm]{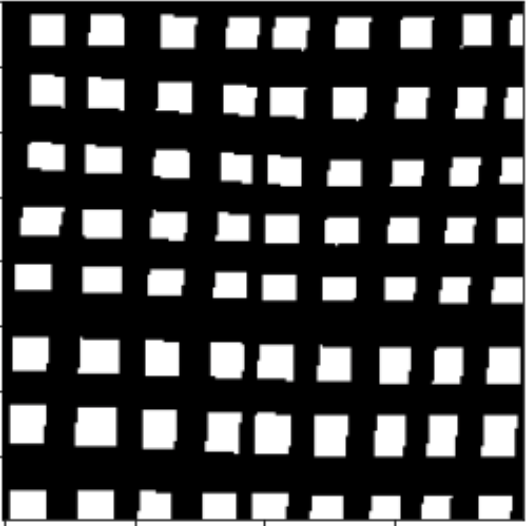}& \includegraphics[width=2.3cm]{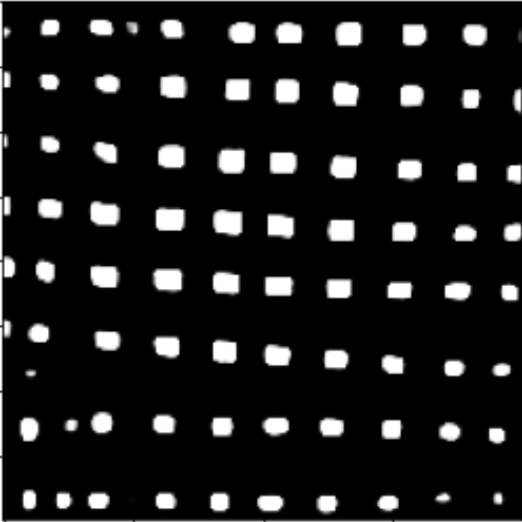}\\
 (d)&(e)&(f)
\end{tabular}
\caption{The two \new{images} with worst errors in the test set in Fig. \ref{fig:errorDLSC} for the horizontal (a) and vertical (d) densities. In (b) and (e) we have the annotations while in (c) and (f) the output of the DL segmentation.} \label{fig:maxerror}
\end{figure}

\begin{figure*}[htp]
\centering
\input{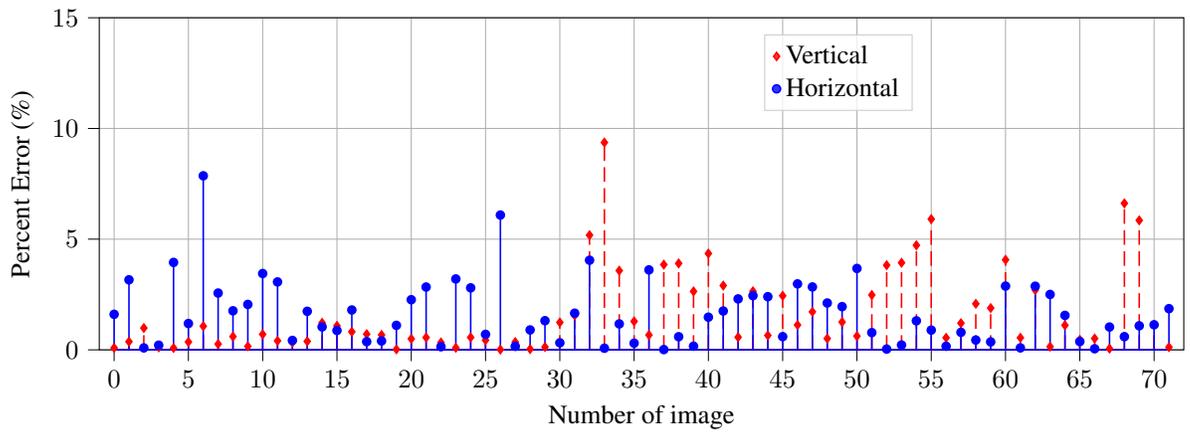}
\caption{Normalized absolute error ($\%$) in horizontal (\textcolor{blue}{$\circ$} solid) and vertical (\textcolor{red}{$\diamond$} dashed) densities for the $1\times 1$ cm corners of the $1.5\times 1.5$ cm annotated \new{samples} in the test set,  DL and SC with $q=10\%$ was used. } \label{fig:errorDLSC}
\end{figure*}

We evaluate the selected model, DLSC, for the test set. The averaged normalized absolute value for test was $1.83\%$ with $q=0\%$ and $1.61\%$ if SC with $q=10\%$ was used. This last value was the one selected for the processing of the whole X-ray of canvases. In Fig. \ref{fig:errorDLSC} we include the error of this DLSC model for \new{images} in the test set. We show the absolute normalized error for the samples in the four $1\times1$ cm corners of the annotated $1.5\times 1.5$ cm \new{samples} in the test set, both for the horizontal and vertical threads. In Fig. \ref{fig:maxerror} we include the input (first column), the labeling (second column) and the output of the DL model (third column) results for \new{images} number 6 (Fig.  \ref{fig:maxerror}.(a)-(c)) and 32 (Fig. \ref{fig:maxerror}.(d)-(f)) \new{in Fig. \ref{fig:errorDLSC}}, the ones with the largest errors in the horizontal and vertical densities, respectively. In the first case, the very thin treads are hard to segment\RED{,} providing a value of $17.67$ thr/cm when the true value was $19.18$ thr/cm. In the second scenario, I) the poor quality of the image and II) the error in the labeling where some \new{crossing} points are missing \new{cause} this error, estimating $9.36$ thr/cm from the segmentation where $8.9$ thr/cm is estimated from the annotations. See the left lower and upper parts in Fig. \ref{fig:maxerror}.(d)-(f) as examples of I) and II), respectively.

\begin{figure*}[htp]
\centering
\input{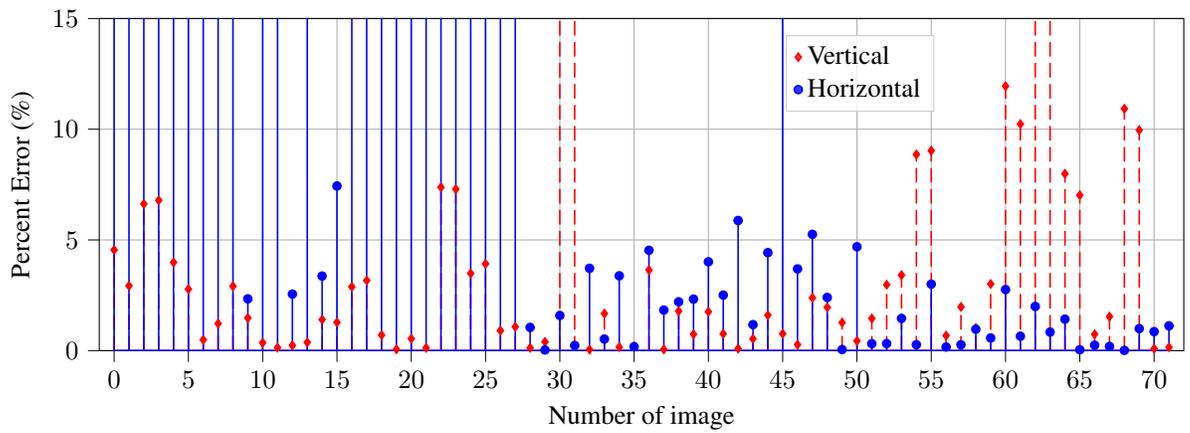}
\caption{Normalized absolute error ($\%$) in horizontal (\textcolor{blue}{$\circ$} solid) and vertical (\textcolor{red}{$\diamond$} dashed) densities  for the $1\times 1$ cm corners of the $1.5\times 1.5$ cm annotated \new{samples} in the test set and the FT approach. } \label{fig:errorFT}
\end{figure*}

In Fig. \ref{fig:errorFT} we include the results for the FT approach applied to the same test \new{images} using 2048 points for the discrete FT. It can be observed how the results for the first 28 \new{images} are quite poor, normalized errors rise up to $50\%$\new{,} being more relevant for the horizontal threads, as vertical ones are better observed in the \new{images}. The overall average normalized error is $7.47\%$. These first 28 \new{images} came from the `The Crucified Christ' by Velázquez (P001167 in MNP inventory) \cite{P001167} where we have a similar aspect to the one in Fig. \ref{fig:maxerror}.(a), i.e.\RED{,} high thread densities and low noise, and the horizontal threads \new{cannot be visually identified as lines}. For the rest of the test \new{images}, to the right, both warp and weft are better observed. However, the FT still provides worse results. For example, above \new{image number 28 in Fig. \ref{fig:errorFT} we have eight images} with error above $10\%$ while in the DLSC\new{,  Fig. \ref{fig:maxerror},} we have none in the whole test set. 

\section{Canvas Analysis}\label{sec:exp}

In the following, \new{we propose two case studies, to check for the} correspondence between pairs of fabrics \new{by using the DLSC approach}. 
We will also include the results of the \new{FT method} as described in \cite{Simois18}. \new{We process $1\times 1$ cm patches in the full preprocessed image, at $200$ ppc, of the X-ray of the canvas, $\ma{Z}$, whose top left corners are at $\ma{Z}[p\cdot s,q\cdot s]$ where $p$ and $q$ are non-negative integers and $s$ is the shift, in pixels, from one patch to the following one}. 

\new{\subsection{Velázquez's Portraits}}

\begin{figure}[htp]
\centering
\begin{tabular}{cccc}
 \includegraphics[width=3.57cm]{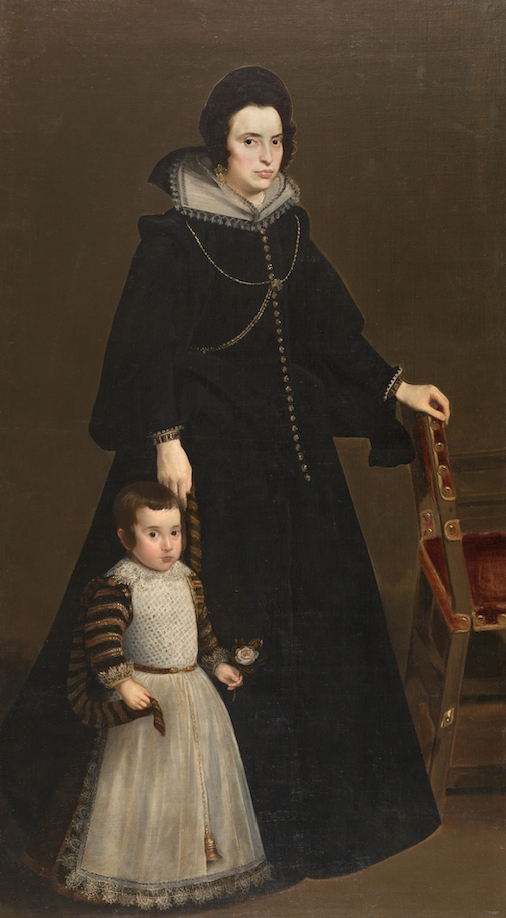}& 
  \includegraphics[width=3.61cm]{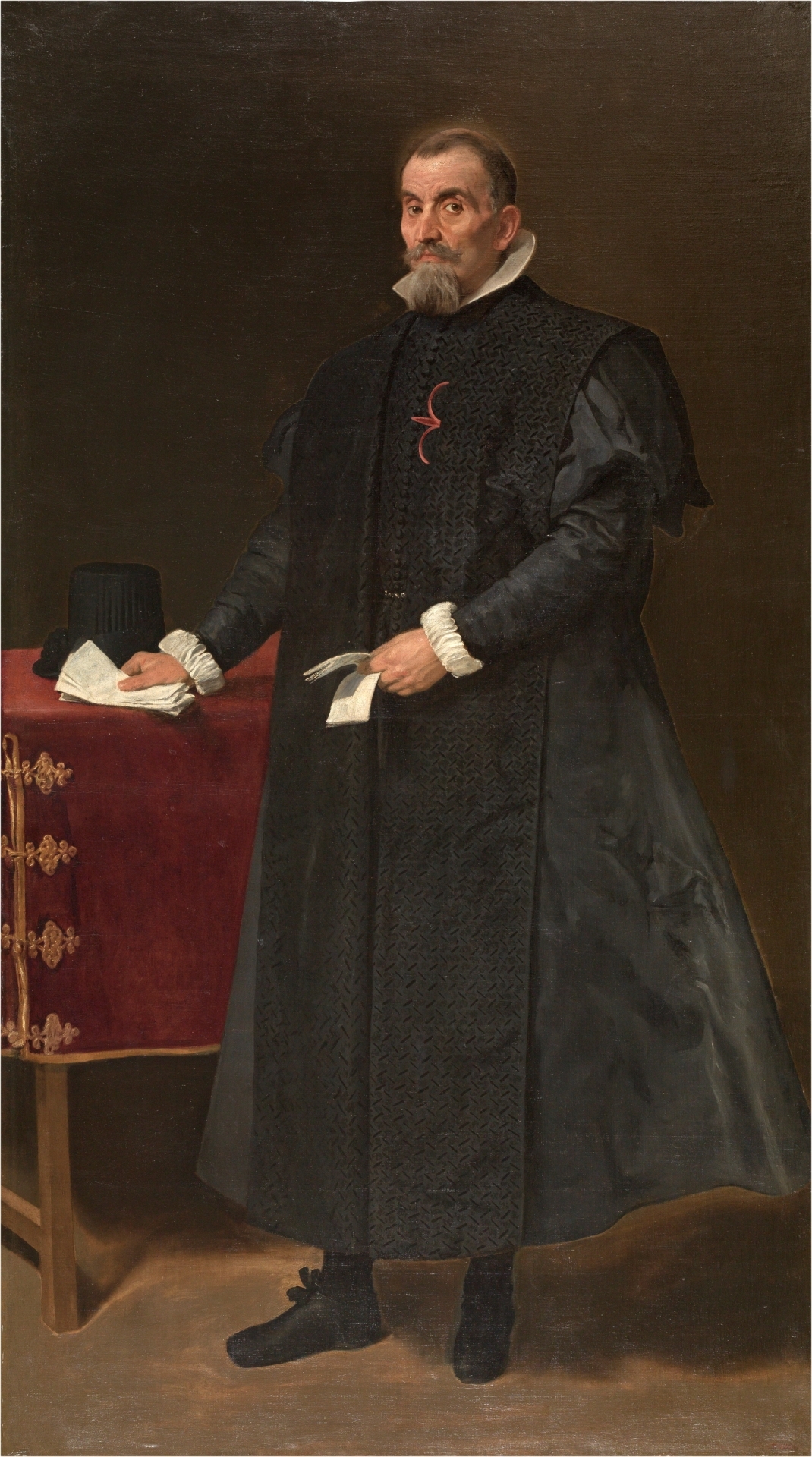}\\
(a) & (b) 
\end{tabular}
\caption{Paintings by Velázquez (a) Antonia de Ipeñarrieta y Galdós and her Son, Luis \cite{P001196} and (b) Diego del Corral y Arellano \cite{P001195}.} \label{fig:P0119X}
\end{figure}

We analyze two canvases by Velázquez, on the one hand Antonia de Ipeñarrieta y Galdós and her Son, Luis (P001196) \cite{P001196} and on the other  Diego del Corral y Arellano (P001195) \cite{P001195}. In this couple of canvases husband and wife were portrayed, see Fig. \ref{fig:P0119X}, and hence it is conjectured that both were painted at the same time on fabrics from the same roll.

\begin{figure}[htp]
\centering
 \includegraphics[width=8.1cm]{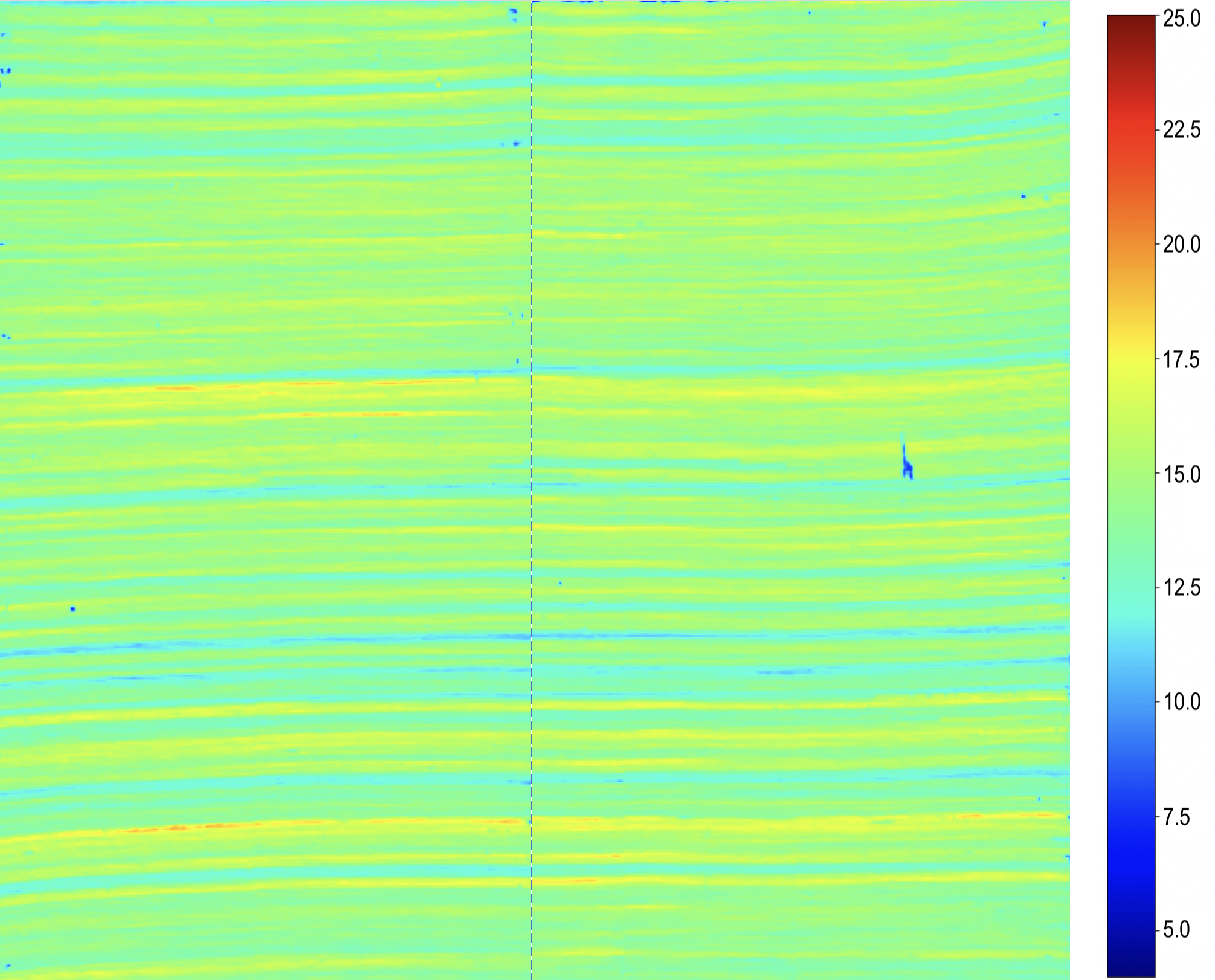}
\caption{Match of horizontal thread densities using DLSC, in thr/cm, for Antonia de Ipeñarrieta and Son (P001196), to the left and Diego del Corral y Arellano (P001195), flipped horizontally, to the right. The maps of densities are separated by a dashed line. } \label{fig:VelazquezJointver}
\end{figure}

We used the DLSC approach with $q=10\%$. In Fig. \ref{fig:VelazquezJointver} we include the horizontal `density map' for both paintings. This density map is the estimated value for the threads counting along the canvas.  \new{With a shift of $s=100$, we have four pixels in the resulting density map for every square centimeter in the canvas.} The color of any pixel encodes the value of the thread density in thr/cm \new{in the corresponding location, $[ps,qs]$}, see \new{the} color bar to the right. 
 In the figure we observe a vertical dashed line. The image to the left corresponds to the horizontal density map of P001196\RED{,} while the one to the right to P001195 horizontally flipped. It can be clearly observed how the pattern of variations of the separation in the horizontal threads perfectly matches in both canvases. This indicates that both fabrics came from the same roll.

\begin{figure}[htp]
\centering
 \includegraphics[width=8.1cm]{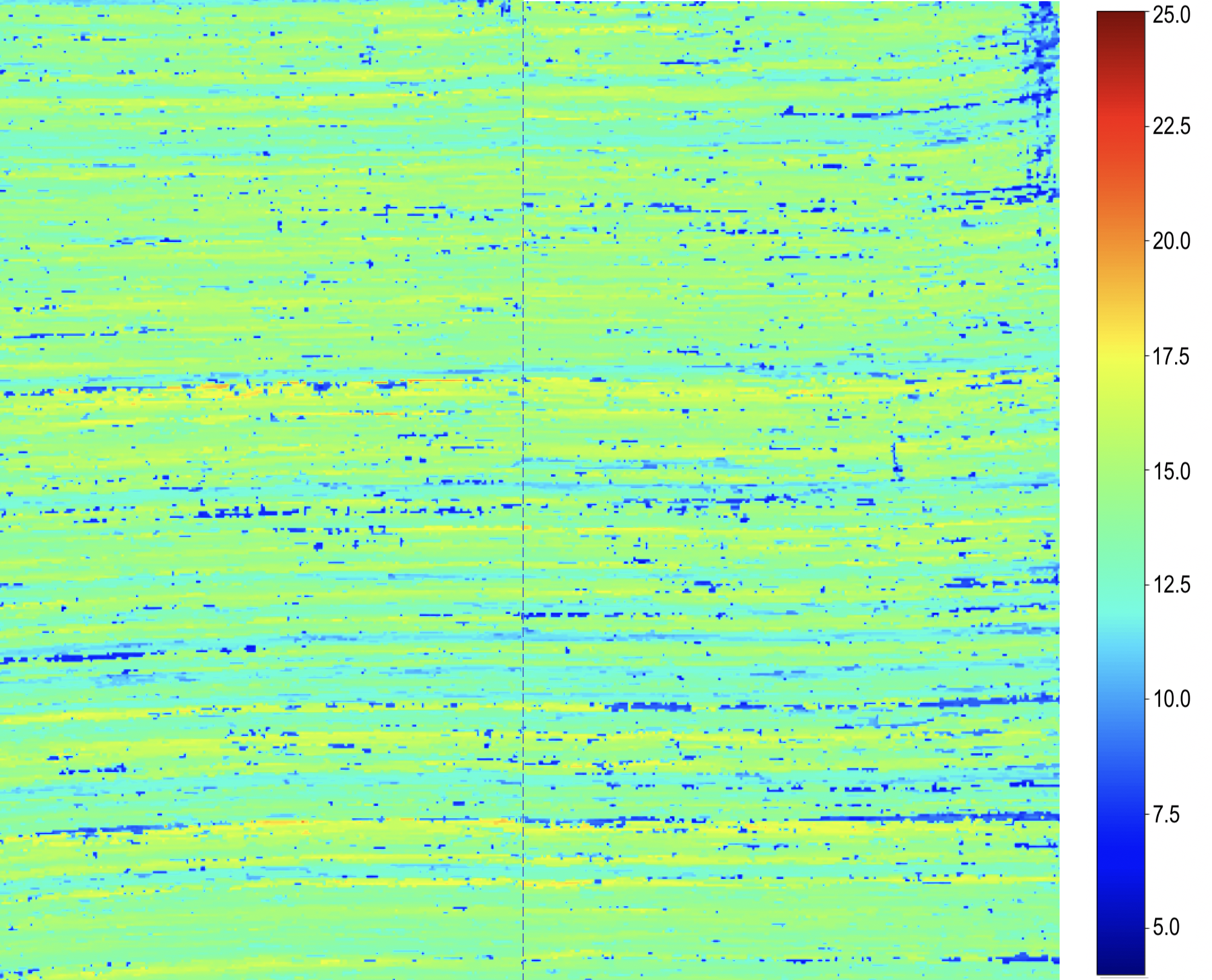}
\caption{Match of horizontal thread densities using FT, in thr/cm, for Antonia de Ipeñarrieta and Son (P001196), to the left and Diego del Corral y Arellano (P001195), flipped horizontally, to the right. The maps of densities are separated by a dashed line.} \label{fig:VelazquezJointFTver}
\end{figure}

\begin{figure}[htp]
\centering
\begin{tabular}{cccc}
   \includegraphics[width=4.22cm]{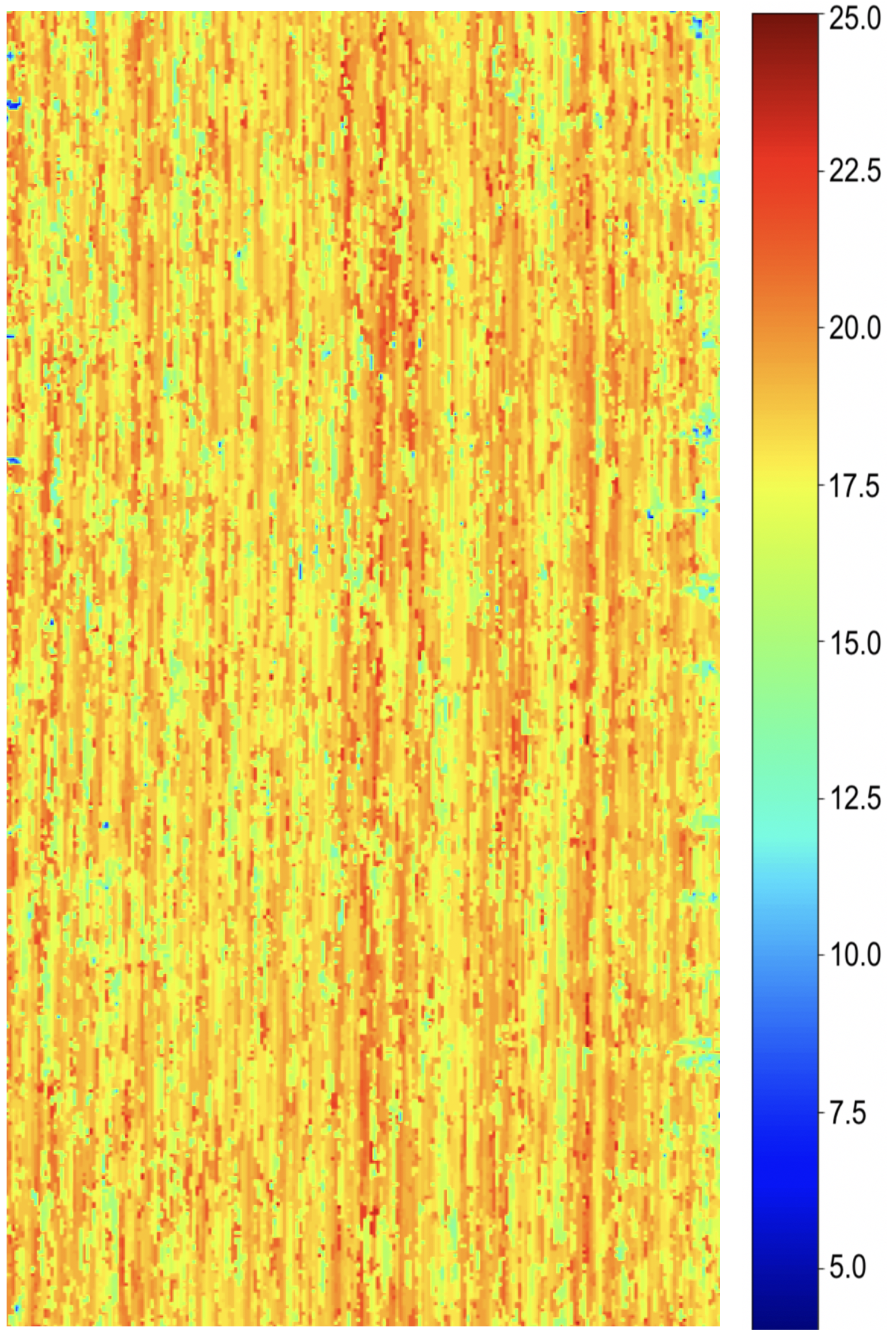}& 
\includegraphics[width=3.4cm]{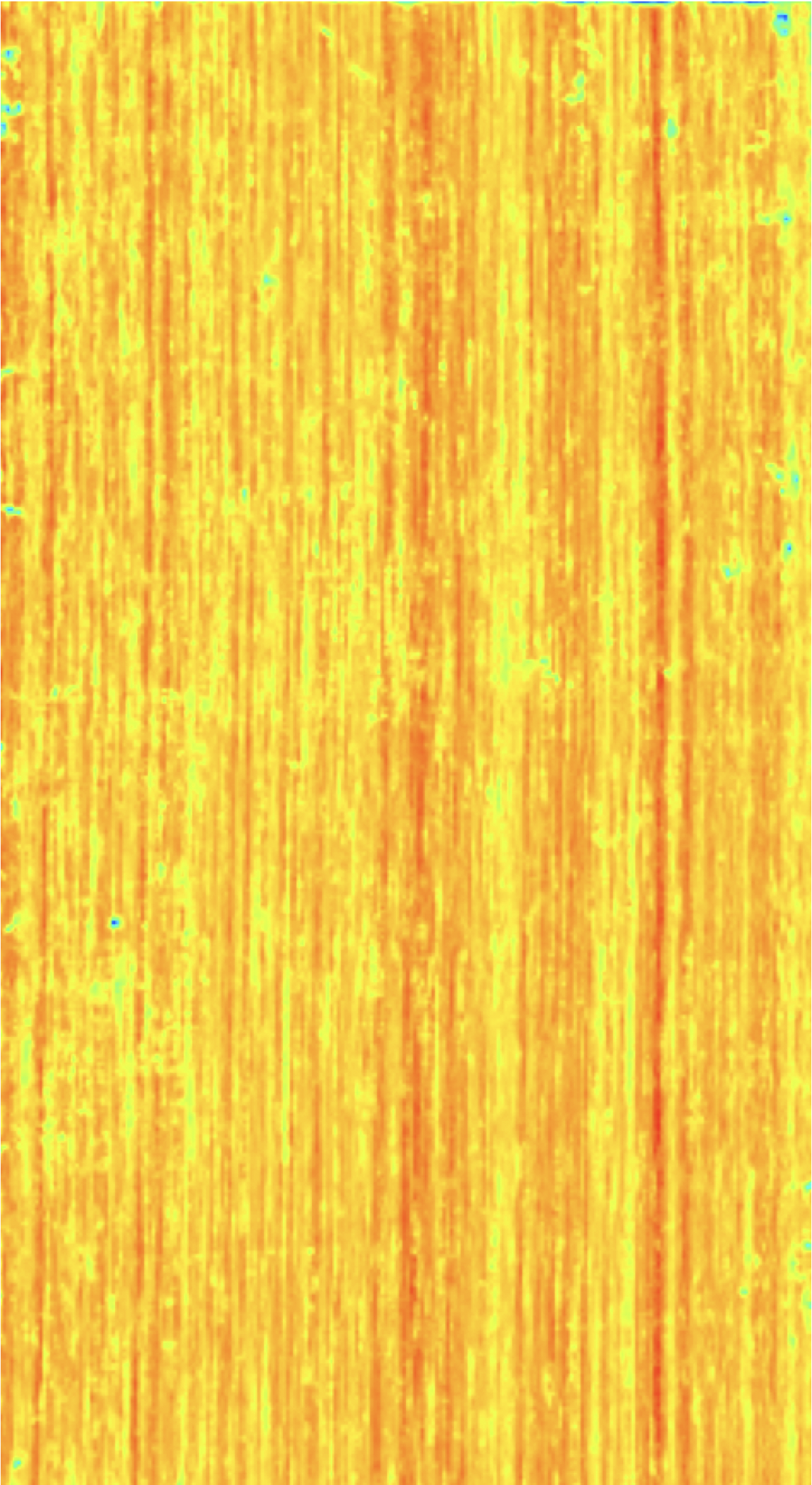}\\
(a) & (b) 
\end{tabular}
\caption{Vertical thread densities, in thr/cm, for Antonia de Ipeñarrieta and Son (P001196) with (a) FT and (b) DLSC.} \label{fig:P01196ver}
\end{figure}

\begin{figure}[htp]
\centering
\begin{tabular}{cccc}
 \includegraphics[width=3.2cm]{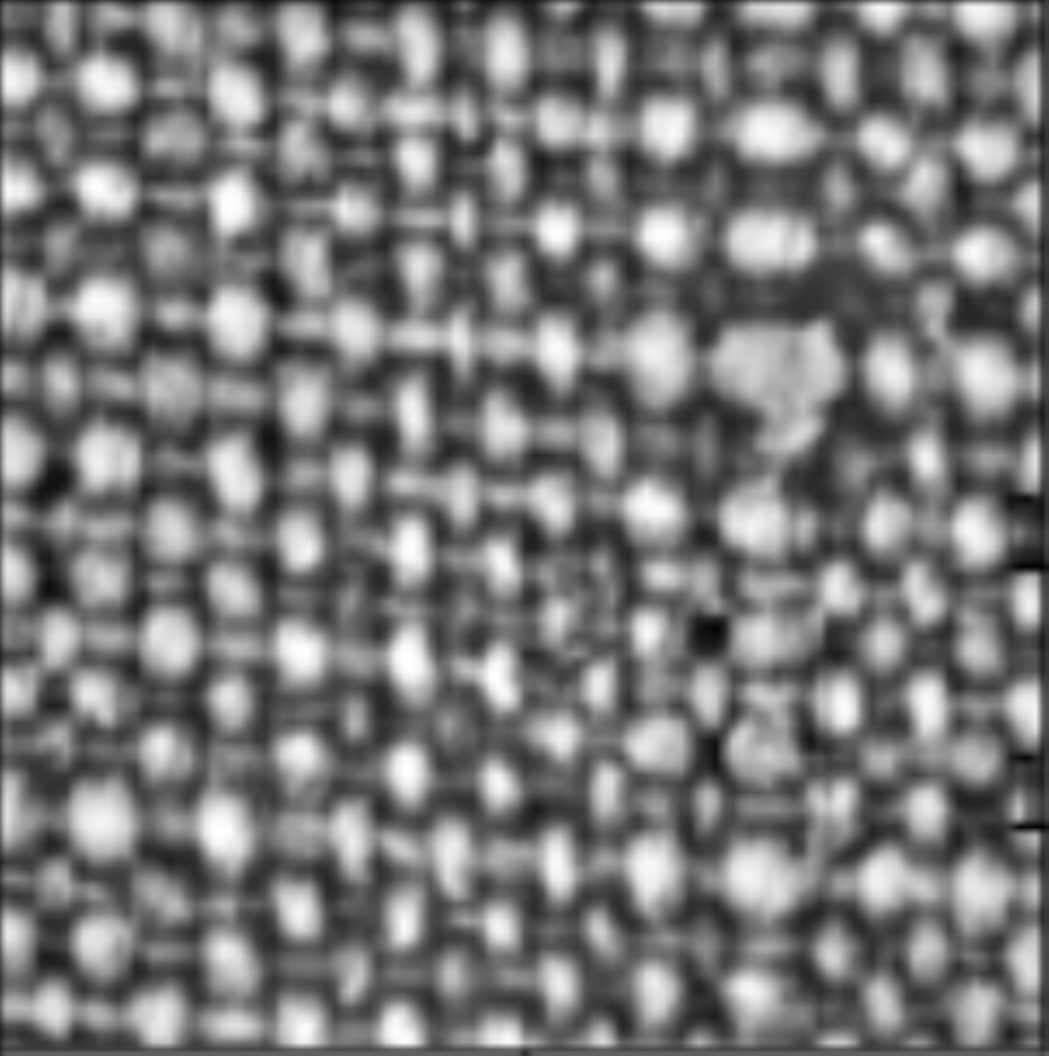}& 
  \includegraphics[width=3.2cm]{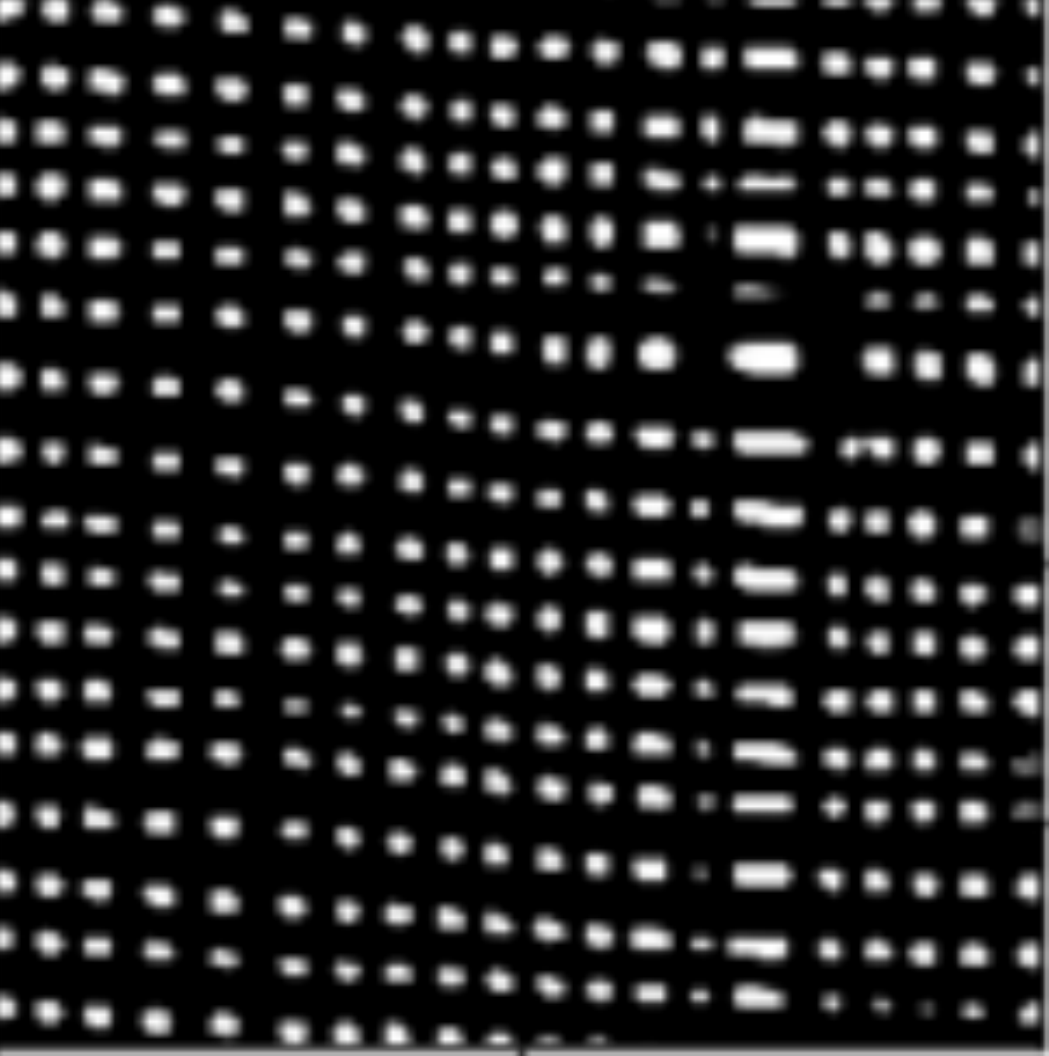}\\
  (a) & (b) \\
   \includegraphics[width=3.2cm]{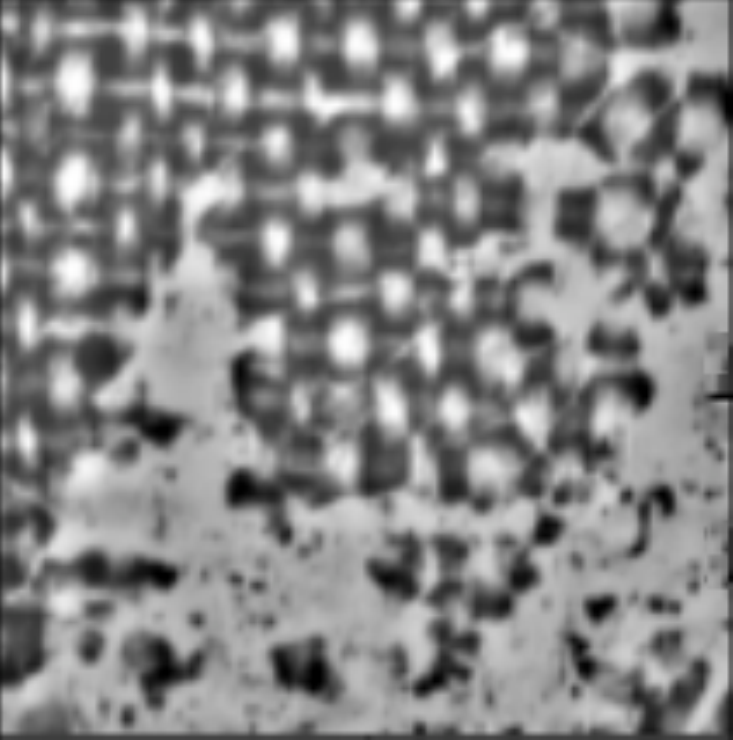}& 
  \includegraphics[width=3.2cm]{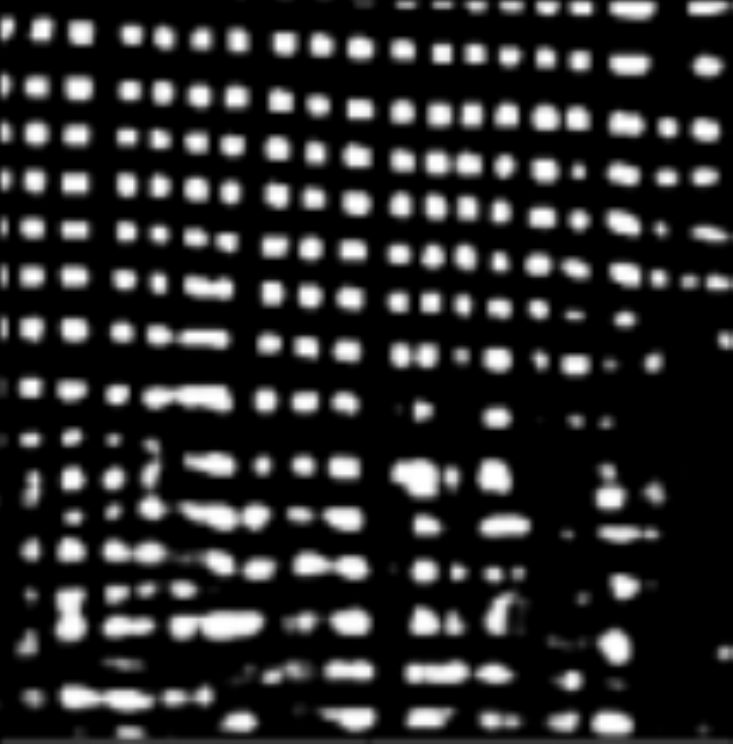}\\
(c) & (d) 
\end{tabular}
\caption{In (a) and (c) \new{patches} of $1\times 1$ cm of the X-ray plate of Diego del Corral y Arellano \new{(P0011195)} and in (b) and (d) the corresponding outputs of the DL based segmentation.} \label{fig:P01195crop}
\end{figure}

In Fig. \ref{fig:VelazquezJointFTver} the same analysis, performed with FT \cite{Simois18}, is included. It can be observed that the result is quite noisy and conclusions about fabric pairing are harder to draw. In Fig. \ref{fig:P01196ver}.(a) and (b) we include the vertical density map for P001196 computed with FT and DLSC respectively. It can be observed that in the vertical density the noise if stronger when using FT, with a very high variability in the vertical axis, while in the DLSC case the result is quite more consistent with the fact that along any vertical line density should not significantly change. To illustrate the differences in the results of the new proposed approach and the FT we \new{focus on} two of the processed \new{patches}, in Fig. \ref{fig:P01195crop}.(a) and (c). In Fig. \ref{fig:P01195crop}.(b) and (d) we have the corresponding outputs of the DL segmentation model. The estimated densities values are analyzed in Tab. \ref{tab:tbl1} for both samples and for the horizontal and vertical threads. Note that since these samples are of size $1\times 1$ cm, the \new{visual} counting of horizontal and vertical threads provides the densities in thr/cm. The densities estimated by FT are quite poor in the case of the first sample, in Fig. \ref{fig:P01195crop}.(a). The main reason, as pointed in Sec. \ref{sec:intro}, is that we have threads with different widths and that in this case vertical threads are poorly observed compared to the horizontal ones. Put in other words, we have both effects described through Fig. \ref{fig:Problems}. \new{This is the reason why}\RED{,} in this scenario, the main frequency found by the FT approach does not \new{correspond} to the thread counting. In Tab. \ref{tab:tbl1} we have an estimation of 16.88 and 19.50 thr/cm for horizontal and vertical threads, respectively, by means of the DLSC approach, while the FT is providing 11.74 and 12.89 thr/cm, i.e.\RED{,} it is focusing more on the thick threads than on the thin ones. By using the DLSC\RED{,} we avoid focusing on main frequencies to measure distances between any pair of consecutive threads. When the segmentation fails, for example because of the poor quality of the image, the estimation of densities through the average distances between threads is not accurate. In Fig. \ref{fig:P01195crop}.(c) the fabric cannot be observed in a large part of the \new{patch}. In the output of the segmentation, Fig. \ref{fig:P01195crop}.(d), we observe that the crossing points are missing in some parts. In the lower part of the output, the estimation of the vertical thread density by using SC introduces an error in the overall estimation, as observed in Tab. \ref{tab:tbl1}. Still, the DLSC is providing a much better estimation than the FT. However, if in this case the FA is used instead \RED{of} the SC, we observe \RED{an} improvement, estimating $19.14$ thr/cm compared to $17.37$ thr/cm of the DLSC and $10.54$ thr/cm of the FT. Usually, we observe outcomes of the DLFA better than the FT but worse than the DLSC while at some points, as in this \new{patch}, the DLFA provides better estimations. 

\begin{table}[htb]
\caption{Estimation of densities for the horizontal and vertical threads in thr/cm for the patches in Fig. \ref{fig:P01195crop} with DL based crossing points segmentation and SC (DLSC), DL based crossing points segmentation with FA (DLFA) and FT approach (FT).}\label{tab:tbl1}
\begin{tabular}{c c c c c c c}
\toprule
& \multicolumn{3}{L}{Horizontal (thr/cm)} & \multicolumn{3}{L}{Vertical (thr/cm)} \\
\new{Sample}  & DLSC& DLFA  & FT &DLSC & DLFA & FT \\
\midrule
Fig. \ref{fig:P01195crop}.(a) & 16.88 & 16.83 & 11.74 & 19.50 & 17.19 & 12.89\\
Fig. \ref{fig:P01195crop}.(c) & 14.47 & 14.50 & 14.47 & 17.37 &19.14& 10.54\\
\bottomrule
\end{tabular}
\end{table}

\begin{figure}[htp]
\centering
\begin{tabular}{cccc}
 \includegraphics[width=3.3cm]{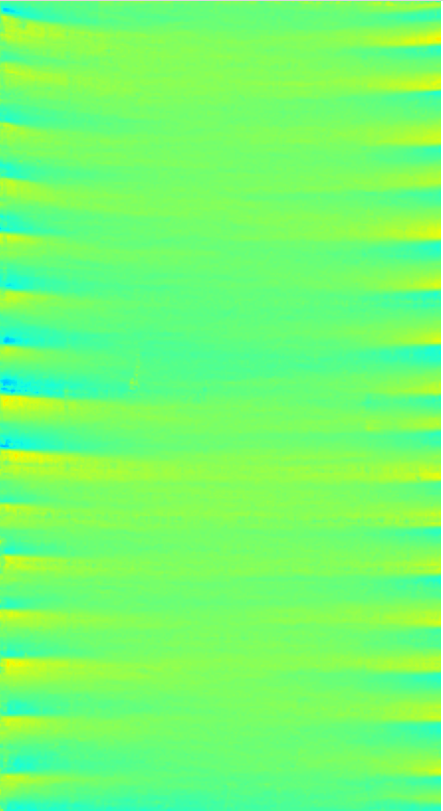}& 
  \includegraphics[width=4.1cm]{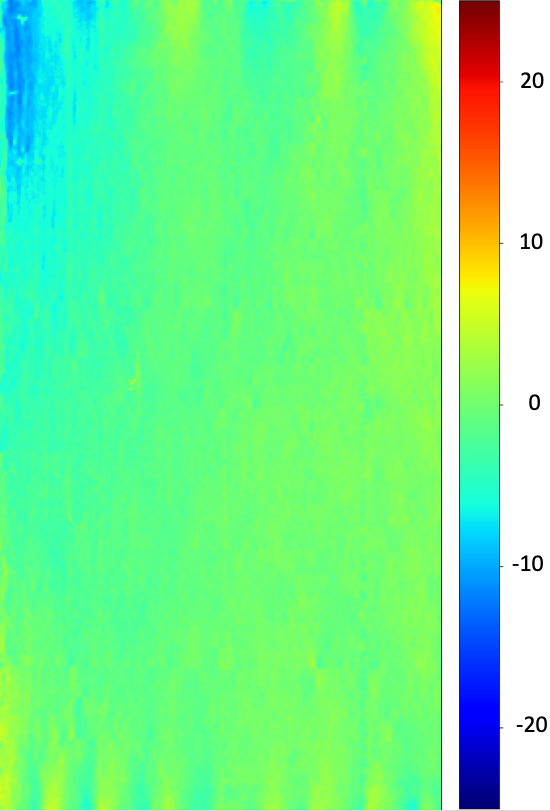}\\
(a) & (b) 
\end{tabular}
\caption{Angle deviation, in degrees, for Diego del Corral y Arellano, of (a) horizontal and (b) vertical threads.} \label{fig:P01195ang}
\end{figure}

Finally, in Fig. \ref{fig:P01195ang}.(a) and (b) we include the angle deviation estimations for P001195 of horizontal and vertical threads, respectively. It can be observed the `garland' effect, i.e., the periodic variation of deviations near the sides of the canvas, due to the separation between nails. This indicates that the painting is conserved in its original size. Also, in the left upper corner of Fig. \ref{fig:P01195ang}.(b) it can be observed that the fabric is twisted. This is consistent with the deformation of lines of thread densities in Fig. \ref{fig:VelazquezJointver}, right upper corner.

\new{The inference in this case took approximately 43 min for each canvas, in the GPU Tesla P100 - CPU Intel Xeon HW. The time to perform inference depends on the number of patches to be processed. As already mentioned we used a 50\% overlap, since the canvas is $2.12\times1.21$ m size, we run the Inc-Dice model plus the spatial counting approach for $212\times121\times4 = 102608$ patches. }

\new{
\subsection{Murillo's Prodigal Son}
In the studies of series of canvases the thread density analysis proves to be useful to check for the use of the same fabric in the paintings. We bring here two masterpieces produced by Murillo for the series illustrating the biblical parable of the prodigal son: ``The prodigal son taking leave of his home'' (P00998) \cite{P00998} and ``The prodigal son squandering his inheritance'' (P00999) \cite{P00999}, see Fig. \ref{fig:MurilloMatch}. In Fig. \ref{fig:MurilloMatch} we also include to the right the results of the vertical thread density estimation with $s=20$ using DLSC, where P00999 has been rotated 180 degrees. It can be observed the good match between the fabrics. In Fig. \ref{fig:P00999} results for the FT and DLSC applied to the canvas P00999 can be compared both for the horizontal (top) and vertical (bottom) thread densities. It is interesting to note the improved definition of the DLSC approach, better allowing for the identification of color lines, representing densities along the warp and weft.

\begin{figure}[htp]
\centering
   \includegraphics[width=.8\linewidth]{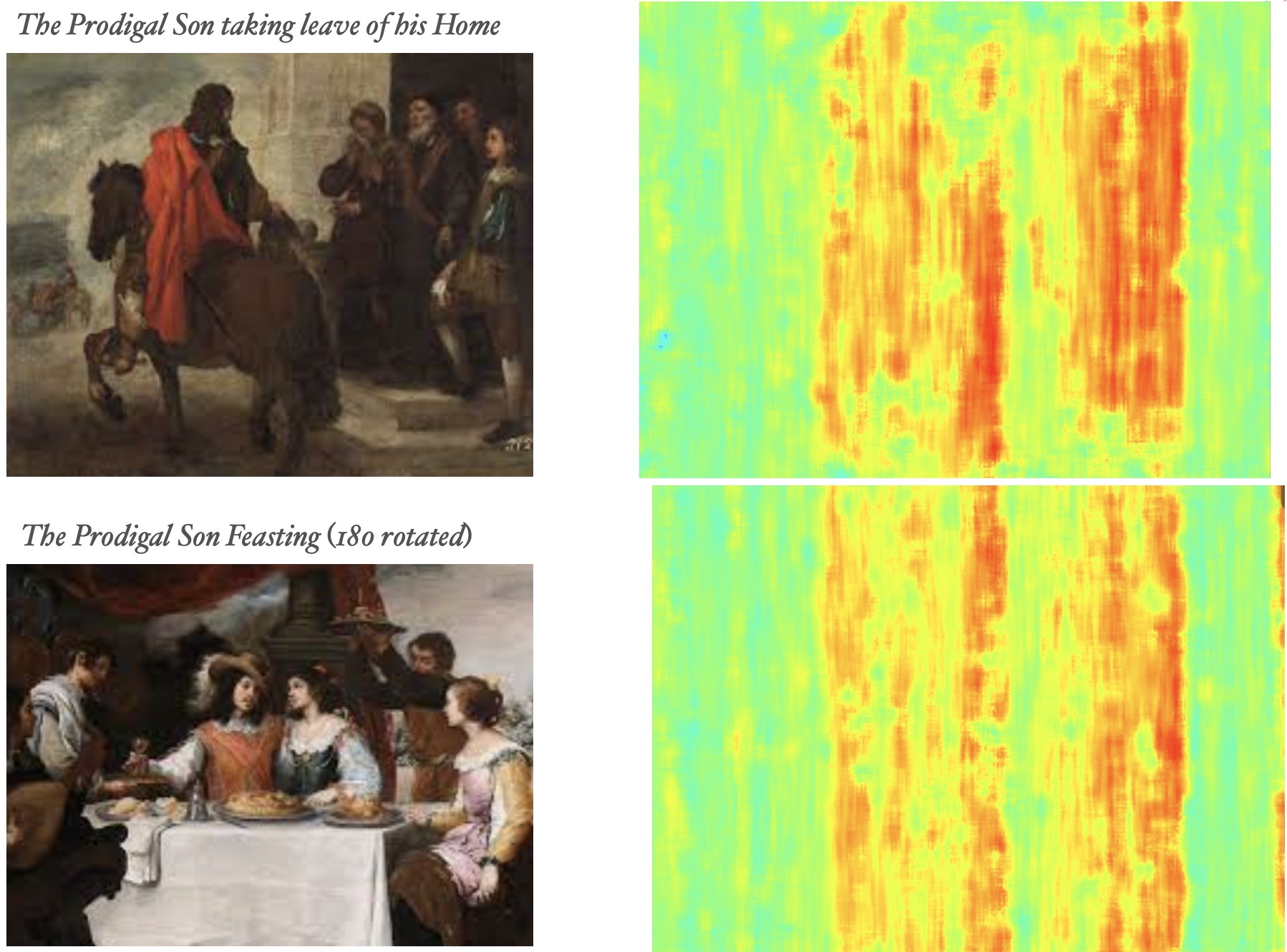}
\caption{Canvases (left) and vertical thread densities (right), in thr/cm, for ``The prodigal son taking leave of his home'' P00998 (top)  and ``The prodigal son squandering his inheritance'' P00999 (bottom, rotated $180^o$).} \label{fig:MurilloMatch}
\end{figure}

\begin{figure}[htp]
\centering
\begin{tabular}{cccc}
   \includegraphics[width=5.697cm]{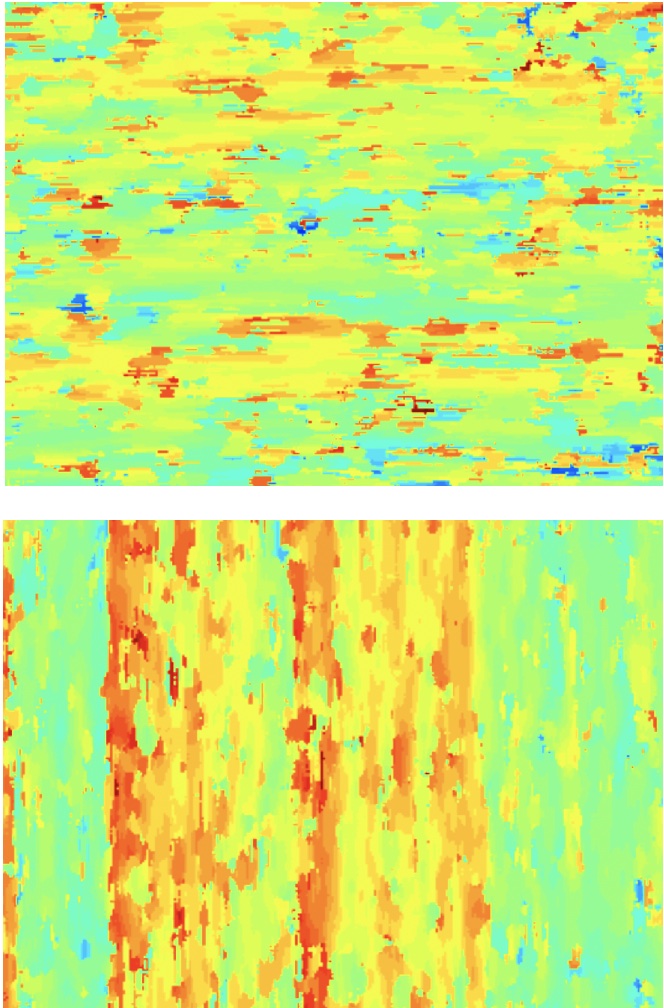}& 
   \includegraphics[width=7.4925cm]{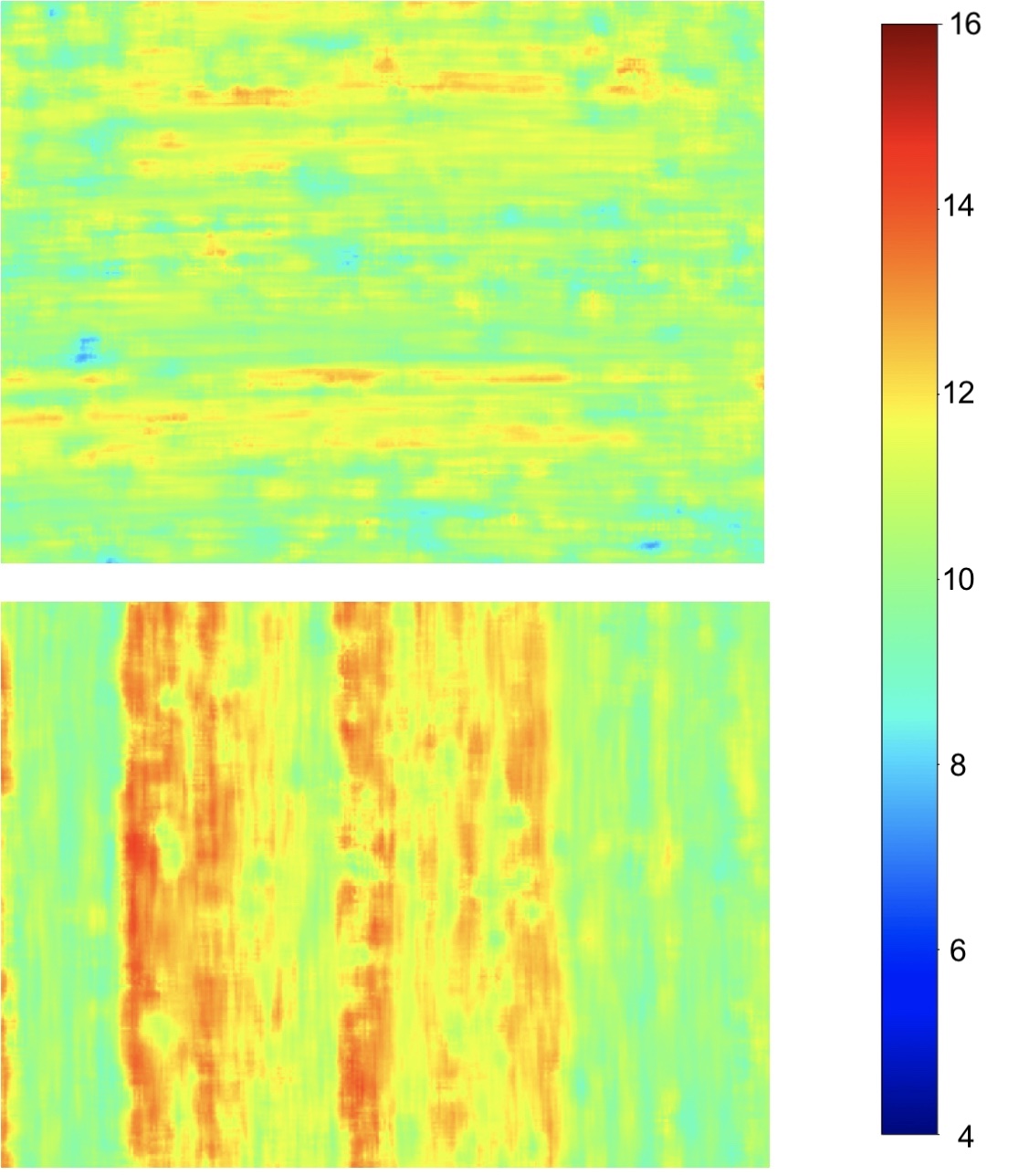}\\
(a) & (b) 
\end{tabular}
\caption{Horizontal (top) and vertical (bottom) thread densities, in thr/cm, for P00999 with (a) FT and (b) DLSC.} \label{fig:P00999}
\end{figure}

}

%

\new{
\subsection{High intensity and low contrast areas}
In Fig. \ref{fig:P01114} we include the results for the estimation of the densities of horizontal threads for one of the X-ray plates of the canvas Ixion by Ribera \cite{P01114}. 
The processed area within the whole canvas is highlighted in Fig. \ref{fig:P01114}.(a) while a $1$ cm side patch is included in Fig. \ref{fig:P01114}.(b). The results for the FT and \RED{DLSC} are reported in Fig. \ref{fig:P01114}.(c) and Fig. \ref{fig:P01114}.(d), respectively. It can be observed that the result of the DLSC is not as good as the one of the FT. Note that the fabric, see Fig. \ref{fig:P01114}.(b), has a low density of threads with a very regular distance between them. In this scenario the FT is quite robust and provides excellent results. On the contrary, the DLSC fails in areas of the plates where we have very high levels and low contrast, because threads cannot be observed in the image. Please pay attention to the shoulder of the man on top and the high intensity area below its arm, within the processed area. This degrades the outcome of the segmentation, and the SC is unable to provide an accurate result as it needs a good enough grid of crossing points. However, a method searching for main frequencies after the segmentation successes, see the outcome of the DLFA in Fig. \ref{fig:P01114}.(e). It can be observed that we achieve similar results to the ones in Fig. \ref{fig:P01114}.(c), with the FT. But in these areas, the DL will not improve the FT. 

\begin{figure}[htp]
\centering

\begin{tabular}{ccccc }
\begin{tabular}{c}
\includegraphics[valign=m,width=5cm]{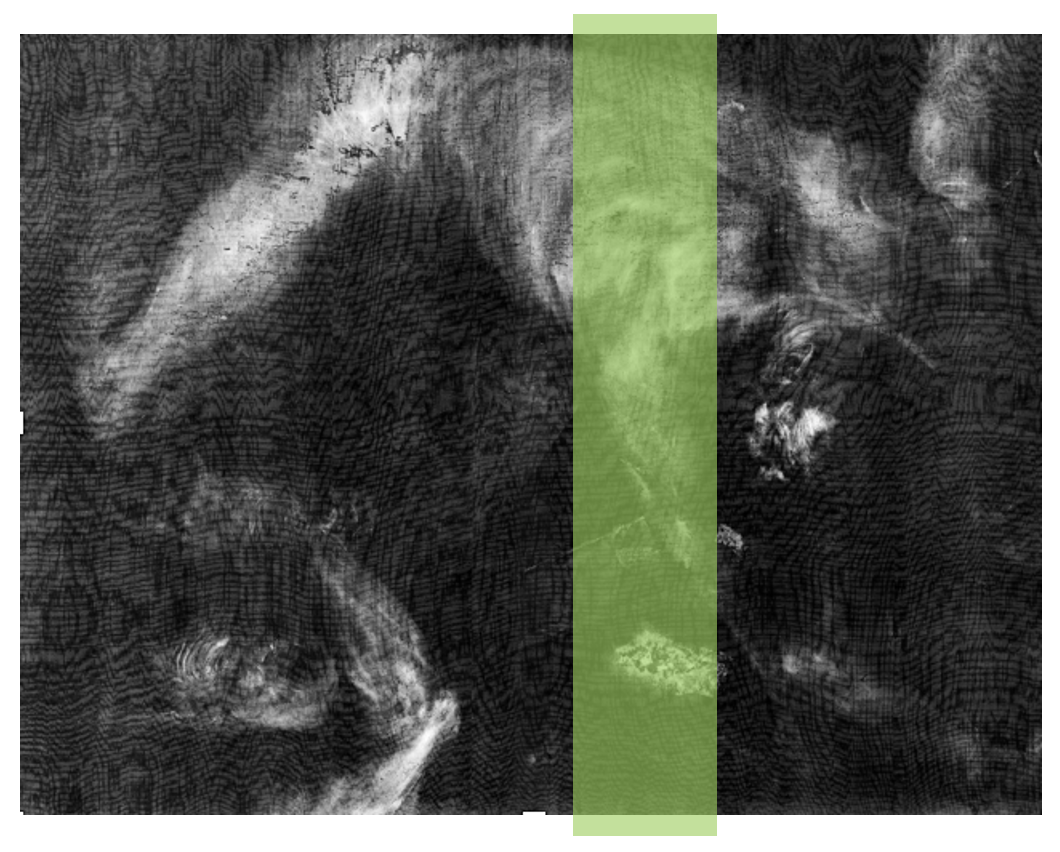}\\(a)\\
\includegraphics[valign=m,width=3cm]{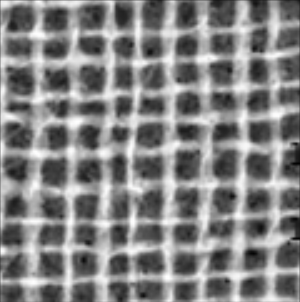}\\(b)
\end{tabular}&
      \includegraphics[valign=m,width=1.5cm]{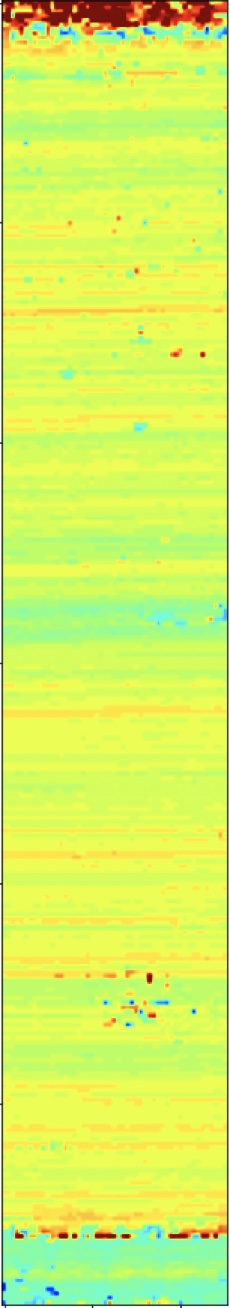}& 
   \includegraphics[valign=m,width=1.48cm]{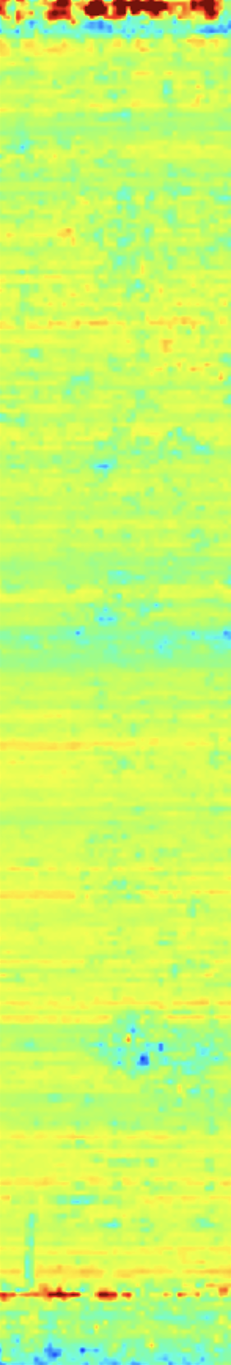}& 
\includegraphics[valign=m,width=1.5cm]{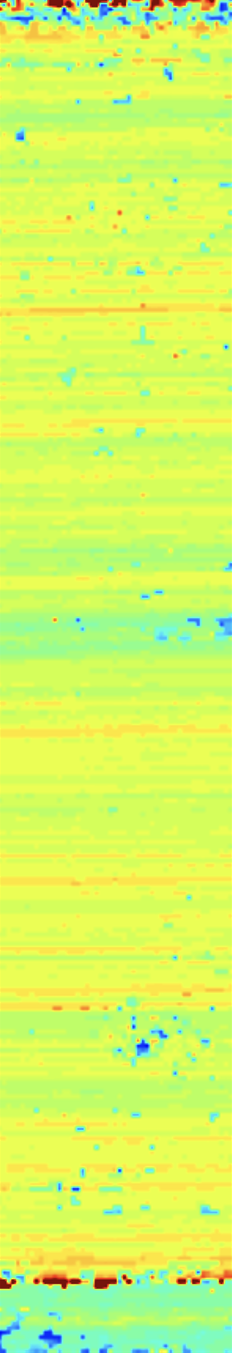}& 
\includegraphics[valign=m,width=0.75cm]{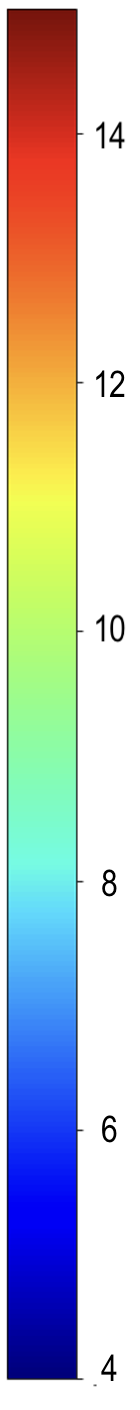}\\
 & (c) & (d) & (e) &
\end{tabular}
\caption{Horizontal thread densities, in thr/cm, for the vertical area highlighted in (a) over the full Xray of Ixion by Ribera \cite{P01114} with (c) FT, (d) DLSC and (e) DLFA. A $1$ cm side patch is included in (b).} \label{fig:P01114}
\end{figure}

}

\section{Conclusions}
\new{
In this paper we present a multidisciplinary investigation in which an effort has been made to understand the problem posed, analyze the state of the art, its drawbacks and study the possible alternatives.
In this process we identify and report two scenarios in which methods based on frequency domain analysis fail. To overcome these problems\RED{,} we propose to resort to the spatial domain by segmenting the crossing points.

The previous known solution working in the spatial domain needs a previous partial labeling of the image to be analyzed. We propose a new algorithm that can use the curator with minimal effort and knowledge, without labeling or a complex parameter selection.

The U-Net, used as starting point, was not capable of processing fabrics with different thread densities, typically in the 5-25 range. Fixed width kernels did not give good results in all cases. We investigated several models with different layers and kernel sizes to conclude that inception was a good option to accommodate the different thread frequencies possible.
We also found that the loss functions used in training had a relevant impact on the result. Consequently, we studied the different possibilities and adopted Dice as the criterion to evaluate the error of the model. 

The generation of the dataset from scratch, another contribution of this work, is also developed. This includes the preprocessing and the data augmentation stages. Labeling of the data is cumbersome. To gain in efficiency\RED{,} we decided to label larger samples (1.5 times wider and taller) than needed at the input of the model, to then use different areas in the data augmentation step. 
The organization of the dataset into training, validation and test was another problem to be addressed. At the beginning of our research we included samples of the same painting in the all of them. Although this has the advantage of having fairly representative subsets, it does not guarantee good performance when a new canvas is presented. It must be taken into account that canvases from different authors may present different counts, different qualities of fabrics or different primers used. 
We designed a partition to ensure that 1) canvases of different thread densities are included in the subsets and 2) different qualities are considered in the training, validation and testing.
In addition, since a small number of instances with very high and very low thread densities are available, in the training phase we had to balance the number of samples used from each image to avoid skewness. 

The proposed deep learning model, after training, provides the crossing points found in the fabric, but we needed to translate the result into the horizontal and vertical counts. We presented an approach to cope with this, the SC, that also provides the estimation of the angle deviations of the threads.
Then we checked the performance at the output of the SC, to observe how a result in segmentation translated into thread densities estimations. Our model selection relied on the final count performance, as discussed in Section \ref{ssec:modelsel}. In Section \ref{ssec:test} we carefully analyzed the final outcome for the patches in the test subset, coming from three new canvases of different densities. In the test set, the whole DLSC procedure, segmentation plus spatial counting, exhibited a very low normalized absolute error, $1.61\%$, while the FT approach had $7.47\%$.

To illustrate the good performance of the novel approach\RED{,} we present two case studies. On the one hand\RED{,} we include a detailed analysis of the canvases P01195 \cite{P001195} and P01196 \cite{P001196} by Velázquez. In this comparison of densities\RED{,} the FT fails to produce a fine enough density estimation maps, as predicted, while the proposed approach provides a very useful result.
We also included a comparison of a pair of canvases of the series The Prodigal Son, by Murillo, to prove that both canvases were painted on canvases coming from the same roll.  Again, our new approach exhibited neater results for the density maps, allowing for a better comparison.

While the DLSC proposed approach presented quite good results in some scenarios, in others its outcome was worse compared to the FT method. It failed for some patches of P001195 by Velázquez or some areas of P001114 by Ribera where the fabric could hardly been observed. In these cases, the DLFA, where frequency analysis was used instead \RED{of} the SC method, could be used.  
%
%
We conjecture that with a better preprocessing stage at the segmentation input we could further improve the DLSC approach. This is a future line of research. 

To sum up,  this novel proposal focuses on the application of deep learning to thread counting in old paintings. This works starts with a careful problem analysis to later face a full design from the dataset design to the spatial thread counting, paying attention to the deep model development and the loss functions used. As a result\RED{,} we have a method that improves the state of the art FT approach in several relevant scenarios, with no need of pre-labeling. 

}

%

%

%
\new{
\appendix
\section{Preprocessing}\label{sec:Prep}
%

The images were enhanced with algorithms based on their local mean and variance when the patches were cropped from the plates to facilitate the labeling and the training of the network. The preprocessing was done in three stages: 
\begin{enumerate}
\item \emph{Local mean filtering.} In the local mean filtering we first compute the average around a pixel then subtract this value to the value of the pixel. The computation of the average value is computed by convolving with a constant value square kernel of size $w=2s+1$. If $\ma{X}$ is the input image and $\ma{X}[i,j]$ is the pixel in the $i$th row and $j$th column, the result of this step is
\begin{equation}
\ma{Y}[i,j] = \ma{X}[i,j] - \frac{1}{w^2}  \sum_{k=i - s}^{i + s} \sum_{l=j - s}^{j + s} \ma{X}[k,l] 
\end{equation}
With the local mean filtering we avoid changes in intensities due to wood stretches or to areas with more opaque paintings. In this work we used $w=13$.

\item \emph{Standard deviation filtering.}
Then, to ensure that the whole range is used we divide by the local standard deviation, avoiding division by zero. We first compute the variance as
\begin{equation}
\sigma_{i,j}^2 ={\frac{1}{w^2} \sum_{k=i - s}^{i + s} \sum_{l=j - s}^{j + s} (\ma{Y}[k,l])^2}
\end{equation} 
where we also used $w=13$. Then the output of this stage is
\begin{equation}
\ma{Z}[i,j] = \ma{Y}[i,j]/\max(\sigma_{i,j},\epsilon)
\end{equation}
where $\epsilon$ is any tiny value.

\item \emph{Clipping and Scaling}. In this step we scale the image, $\ma{Z}$, to use the full dynamic range by setting its lowest and largest values to zero and one, respectively. Prior to this step we clip low probable largest and lowest values as follows. We first estimate the probabilities of values to be into one out of 256 segments of same lengths between the minimum and maximum values. Extreme values with probabilities below some threshold, $\gamma$, are clipped. We used $\gamma = 10^{-3}$.

\end{enumerate}
}


\begin{ack}
This document is the results of the ATENEA Project P20\_01216 research
   project funded by the Consejería de Transformación Económica, Industria, Conocimiento y Universidades, Junta de Andalucía and European Union in the framework of the FEDER Program.\end{ack}

\section*{References}
 \bibliographystyle{elsarticle-num-names}


\begin{thebibliography}{40}
\expandafter\ifx\csname natexlab\endcsname\relax\def\natexlab#1{#1}\fi
\providecommand{\url}[1]{\texttt{#1}}
\providecommand{\href}[2]{#2}
\providecommand{\path}[1]{#1}
\providecommand{\DOIprefix}{doi:}
\providecommand{\ArXivprefix}{arXiv:}
\providecommand{\URLprefix}{URL: }
\providecommand{\Pubmedprefix}{pmid:}
\providecommand{\doi}[1]{\href{http://dx.doi.org/#1}{\path{#1}}}
\providecommand{\Pubmed}[1]{\href{pmid:#1}{\path{#1}}}
\providecommand{\bibinfo}[2]{#2}
\ifx\xfnm\relax \def\xfnm[#1]{\unskip,\space#1}\fi
\bibitem[{Alba and Murillo-Fuentes(2021)}]{Alba21}
\bibinfo{author}{L.~Alba}, \bibinfo{author}{J.~J. Murillo-Fuentes},
\newblock \bibinfo{title}{Fabrics as a painting support. new tools for the
  study},
\newblock in: \bibinfo{booktitle}{{La ciencia y el arte. Ciencias
  experimentales y conservaci{\'o}n del patrimonio}},
  \bibinfo{publisher}{Ministerio de Cultura y Deporte}, \bibinfo{year}{2021},
  pp. \bibinfo{pages}{219--230}. \DOIprefix\doi{10.1007/978-3-319-75316-4_7}.
\bibitem[{de~Carbonnel(1980)}]{Vanderlip80}
\bibinfo{author}{K.~V. de~Carbonnel},
\newblock \bibinfo{title}{A study of french painting canvases},
\newblock \bibinfo{journal}{Journal of the American Institute for Conservation}
  \bibinfo{volume}{20} (\bibinfo{year}{1980}) \bibinfo{pages}{3--20}.
\bibitem[{Johnson et~al.(2013)Johnson, Jr., and Erdmann}]{Johnson2013}
\bibinfo{author}{D.~H. Johnson}, \bibinfo{author}{C.~R.~J. Jr.},
  \bibinfo{author}{R.~G. Erdmann},
\newblock \bibinfo{title}{Weave analysis of paintings on canvas from
  radiographs},
\newblock \bibinfo{journal}{Signal Processing} \bibinfo{volume}{93}
  (\bibinfo{year}{2013}) \bibinfo{pages}{527--540}.
\bibitem[{Simois and Murillo-Fuentes(2018)}]{Simois18}
\bibinfo{author}{F.~J. Simois}, \bibinfo{author}{J.~J. Murillo-Fuentes},
\newblock \bibinfo{title}{On the power spectral density applied to the analysis
  of old canvases},
\newblock \bibinfo{journal}{Signal Processing} \bibinfo{volume}{143}
  (\bibinfo{year}{2018}) \bibinfo{pages}{253--268}.
\bibitem[{Rubens(1628)}]{P001692}
\bibinfo{author}{P.~P. Rubens}, \bibinfo{title}{Andan and {E}va},
  \bibinfo{howpublished}{Museo Nacional del Prado (P001692)},
  \bibinfo{year}{1628}.
\bibitem[{de~Silva~y Vel{\'a}zquez(1634)}]{P001180}
\bibinfo{author}{D.~R. de~Silva~y Vel{\'a}zquez}, \bibinfo{title}{Prince
  {B}altasar {C}arlos on horseback}, \bibinfo{howpublished}{Museo Nacional del
  Prado (P001180)}, \bibinfo{year}{1634}.
\bibitem[{Minaee et~al.(2022)Minaee, Boykov, Porikli, Plaza, Kehtarnavaz, and
  Terzopoulos}]{Shervin2022}
\bibinfo{author}{S.~Minaee}, \bibinfo{author}{Y.~Boykov},
  \bibinfo{author}{F.~Porikli}, \bibinfo{author}{A.~Plaza},
  \bibinfo{author}{N.~Kehtarnavaz}, \bibinfo{author}{D.~Terzopoulos},
\newblock \bibinfo{title}{Image segmentation using deep learning: A survey},
\newblock \bibinfo{journal}{IEEE Transactions on Pattern Analysis and Machine
  Intelligence} \bibinfo{volume}{44} (\bibinfo{year}{2022})
  \bibinfo{pages}{3523--3542}.
\bibitem[{Szegedy et~al.(2015)Szegedy, Liu, Jia, Sermanet, Reed, Anguelov,
  Erhan, Vanhoucke, and Rabinovich}]{Inception14}
\bibinfo{author}{C.~Szegedy}, \bibinfo{author}{W.~Liu},
  \bibinfo{author}{Y.~Jia}, \bibinfo{author}{P.~Sermanet},
  \bibinfo{author}{S.~Reed}, \bibinfo{author}{D.~Anguelov},
  \bibinfo{author}{D.~Erhan}, \bibinfo{author}{V.~Vanhoucke},
  \bibinfo{author}{A.~Rabinovich},
\newblock \bibinfo{title}{Going deeper with convolutions},
\newblock in: \bibinfo{booktitle}{2015 IEEE Conference on Computer Vision and
  Pattern Recognition (CVPR)}, \bibinfo{year}{2015}, pp. \bibinfo{pages}{1--9}.
  \DOIprefix\doi{10.1109/CVPR.2015.7298594}.
\bibitem[{Aradillas et~al.(2021)Aradillas, Murillo-Fuentes, and
  Olmos}]{Aradillas21}
\bibinfo{author}{J.~C. Aradillas}, \bibinfo{author}{J.~J. Murillo-Fuentes},
  \bibinfo{author}{P.~M. Olmos},
\newblock \bibinfo{title}{Boosting offline handwritten text recognition in
  historical documents with few labeled lines},
\newblock \bibinfo{journal}{IEEE Access} \bibinfo{volume}{9}
  (\bibinfo{year}{2021}) \bibinfo{pages}{76674--76688}.
\bibitem[{{Maaten} and Erdmann(2015)}]{Maaten15}
\bibinfo{author}{L.~{Maaten}}, \bibinfo{author}{R.~G. Erdmann},
\newblock \bibinfo{title}{{Automatic thread-level canvas analysis: A
  machine-learning approach to analyzing the canvas of paintings}},
\newblock \bibinfo{journal}{IEEE Signal Process. Mag.}  (\bibinfo{year}{2015}).
\bibitem[{Barni et~al.(2005)Barni, Pelagotti, and Piva}]{Barni05}
\bibinfo{author}{M.~Barni}, \bibinfo{author}{A.~Pelagotti},
  \bibinfo{author}{A.~Piva},
\newblock \bibinfo{title}{{Image processing for the analysis and conservation
  of paintings: Opportunities and challenges}},
\newblock \bibinfo{journal}{IEEE Signal Process. Mag.}  (\bibinfo{year}{2005}).
\bibitem[{Cornelis et~al.(2017)Cornelis, Yang, Goodfriend, Ocon, Lu, and
  Daubechies}]{Cornelis17}
\bibinfo{author}{B.~Cornelis}, \bibinfo{author}{H.~Yang},
  \bibinfo{author}{A.~Goodfriend}, \bibinfo{author}{N.~Ocon},
  \bibinfo{author}{J.~Lu}, \bibinfo{author}{I.~Daubechies},
\newblock \bibinfo{title}{Removal of canvas patterns in digital acquisitions of
  paintings},
\newblock \bibinfo{journal}{IEEE Transactions on Image Processing}
  \bibinfo{volume}{26} (\bibinfo{year}{2017}) \bibinfo{pages}{160 -- 171}.
\bibitem[{Deligiannis et~al.(2017)Deligiannis, Mota, Cornelis, Rodrigues, and
  Daubechies}]{Deligiannis2017}
\bibinfo{author}{N.~Deligiannis}, \bibinfo{author}{J.~F.~C. Mota},
  \bibinfo{author}{B.~Cornelis}, \bibinfo{author}{M.~R.~D. Rodrigues},
  \bibinfo{author}{I.~Daubechies},
\newblock \bibinfo{title}{Multi-modal dictionary learning for image separation
  with application in art investigation},
\newblock \bibinfo{journal}{IEEE Transactions on Image Processing}
  \bibinfo{volume}{26} (\bibinfo{year}{2017}) \bibinfo{pages}{751--764}.
\bibitem[{Johnson et~al.(2008)Johnson, Hendriks, Berezhnoy, Brevdo, Hughes,
  Daubechies, Li, Postma, and Wang}]{Johnson08}
\bibinfo{author}{C.~Johnson}, \bibinfo{author}{E.~Hendriks},
  \bibinfo{author}{I.~Berezhnoy}, \bibinfo{author}{E.~Brevdo},
  \bibinfo{author}{S.~Hughes}, \bibinfo{author}{I.~Daubechies},
  \bibinfo{author}{J.~Li}, \bibinfo{author}{E.~Postma},
  \bibinfo{author}{J.~Wang},
\newblock \bibinfo{title}{{Image processing for artist identification}},
\newblock \bibinfo{journal}{IEEE Signal Process. Mag.}  (\bibinfo{year}{2008}).
\bibitem[{Rucoba-Calder{\'o}n et~al.(2022)Rucoba-Calder{\'o}n, Ramos, and
  Guti{\'e}rrez-C{\'a}rdenas}]{Rucoba22}
\bibinfo{author}{C.~Rucoba-Calder{\'o}n}, \bibinfo{author}{E.~Ramos},
  \bibinfo{author}{J.~Guti{\'e}rrez-C{\'a}rdenas},
\newblock \bibinfo{title}{Crack detection in oil paintings using morphological
  filters and {K-SVD} algorithm},
\newblock in: \bibinfo{editor}{J.~A. Lossio-Ventura},
  \bibinfo{editor}{J.~Valverde-Rebaza}, \bibinfo{editor}{E.~D{\'\i}az},
  \bibinfo{editor}{D.~Mu{\~{n}}ante}, \bibinfo{editor}{C.~Gavidia-Calderon},
  \bibinfo{editor}{A.~D.~B. Valejo}, \bibinfo{editor}{H.~Alatrista-Salas}
  (Eds.), \bibinfo{booktitle}{Information Management and Big Data},
  \bibinfo{publisher}{Springer International Publishing},
  \bibinfo{address}{Cham}, \bibinfo{year}{2022}, pp. \bibinfo{pages}{329--339}.
\bibitem[{Sizyakin et~al.(2020)Sizyakin, Cornelis, Meeus, Dubois, Martens,
  Voronin, and Pizurica}]{Sizyakin20}
\bibinfo{author}{R.~Sizyakin}, \bibinfo{author}{B.~Cornelis},
  \bibinfo{author}{L.~Meeus}, \bibinfo{author}{H.~Dubois},
  \bibinfo{author}{M.~Martens}, \bibinfo{author}{V.~Voronin},
  \bibinfo{author}{A.~Pizurica},
\newblock \bibinfo{title}{Crack detection in paintings using convolutional
  neural networks},
\newblock \bibinfo{journal}{IEEE Access} \bibinfo{volume}{8}
  (\bibinfo{year}{2020}) \bibinfo{pages}{74535 -- 74552}. \bibinfo{note}{Cited
  by: 8; All Open Access, Gold Open Access, Green Open Access}.
\bibitem[{Roberto et~al.(2020)Roberto, Ortego, and Davis}]{Roberto2020}
\bibinfo{author}{J.~Roberto}, \bibinfo{author}{D.~Ortego},
  \bibinfo{author}{B.~Davis},
\newblock \bibinfo{title}{Toward the automatic retrieval and annotation of
  outsider art images: A preliminary statement},
\newblock in: \bibinfo{booktitle}{AI4HI}, \bibinfo{year}{2020}.
\bibitem[{{Pu} et~al.(2020){Pu}, {Sober}, {Daly}, {Higgitt}, {Daubechies}, and
  {Rodrigues}}]{Pu2020}
\bibinfo{author}{W.~{Pu}}, \bibinfo{author}{B.~{Sober}},
  \bibinfo{author}{N.~{Daly}}, \bibinfo{author}{C.~{Higgitt}},
  \bibinfo{author}{I.~{Daubechies}}, \bibinfo{author}{M.~R.~D. {Rodrigues}},
\newblock \bibinfo{title}{A connected auto-encoders based approach for image
  separation with side information: With applications to art investigation},
\newblock in: \bibinfo{booktitle}{IEEE Int. Conf. on Acoustics, Speech and
  Signal Process. (ICASSP)}, \bibinfo{year}{2020}, pp.
  \bibinfo{pages}{2213--2217}.
\bibitem[{Zou et~al.(2021)Zou, Zhao, and Zhao}]{Zou21}
\bibinfo{author}{Z.~Zou}, \bibinfo{author}{P.~Zhao}, \bibinfo{author}{X.~Zhao},
\newblock \bibinfo{title}{Virtual restoration of the colored paintings on
  weathered beams in the forbidden city using multiple deep learning
  algorithms},
\newblock \bibinfo{journal}{Advanced Engineering Informatics}
  \bibinfo{volume}{50} (\bibinfo{year}{2021}). \bibinfo{note}{Cited by: 0}.
\bibitem[{Polatkan et~al.(2009)Polatkan, Jafarpour, Brasoveanu, Hughes, and
  Daubechies}]{Polatkan09}
\bibinfo{author}{G.~Polatkan}, \bibinfo{author}{S.~Jafarpour},
  \bibinfo{author}{A.~Brasoveanu}, \bibinfo{author}{S.~Hughes},
  \bibinfo{author}{I.~Daubechies},
\newblock \bibinfo{title}{Detection of forgery in paintings using supervised
  learning},
\newblock in: \bibinfo{booktitle}{2009 16th IEEE International Conference on
  Image Processing (ICIP)}, \bibinfo{year}{2009}, pp.
  \bibinfo{pages}{2921--2924}. \DOIprefix\doi{10.1109/ICIP.2009.5413338}.
\bibitem[{Nemade et~al.(2017)Nemade, Nitsure, Hirve, and Mane}]{Nemade2017}
\bibinfo{author}{R.~Nemade}, \bibinfo{author}{A.~Nitsure},
  \bibinfo{author}{P.~Hirve}, \bibinfo{author}{S.~B. Mane},
\newblock \bibinfo{title}{Detection of forgery in art paintings using machine
  learning},
\newblock \bibinfo{year}{2017}.
\bibitem[{Ronneberger et~al.(2015)Ronneberger, Fischer, and Brox}]{Unet15}
\bibinfo{author}{O.~Ronneberger}, \bibinfo{author}{P.~Fischer},
  \bibinfo{author}{T.~Brox},
\newblock \bibinfo{title}{{U-N}et: Convolutional networks for biomedical image
  segmentation},
\newblock in: \bibinfo{booktitle}{Lecture Notes in Computer Science (including
  subseries Lecture Notes in Artificial Intelligence and Lecture Notes in
  Bioinformatics)}, \bibinfo{year}{2015}.
  \DOIprefix\doi{10.1007/978-3-319-24574-4_28}.
\bibitem[{Rumelhart et~al.(1986)Rumelhart, Hinton, and Williams}]{AE86}
\bibinfo{author}{D.~Rumelhart}, \bibinfo{author}{G.~Hinton},
  \bibinfo{author}{R.~Williams}, \bibinfo{title}{Neurocomputing: foundations of
  research, learning internal representations by error propagation},
  \bibinfo{year}{1986}.
\bibitem[{Goodfellow et~al.(2016)Goodfellow, Bengio, and
  Courville}]{Goodfellow2016}
\bibinfo{author}{I.~Goodfellow}, \bibinfo{author}{Y.~Bengio},
  \bibinfo{author}{A.~Courville}, \bibinfo{title}{{Deep Learning}},
  \bibinfo{publisher}{MIT Press}, \bibinfo{year}{2016}.
\bibitem[{Escofet et~al.(2001)Escofet, Mill{\'{a}}n, and
  Rall{\'{o}}}]{Escofet2001}
\bibinfo{author}{J.~Escofet}, \bibinfo{author}{M.~S. Mill{\'{a}}n},
  \bibinfo{author}{M.~Rall{\'{o}}},
\newblock \bibinfo{title}{{Modeling of woven fabric structures based on Fourier
  image analysis}},
\newblock \bibinfo{journal}{Applied Optics}  (\bibinfo{year}{2001}).
\bibitem[{Johnson et~al.(2010)Johnson, Sun, Johnson, and
  Hendriks}]{Johnson2010}
\bibinfo{author}{D.~H. Johnson}, \bibinfo{author}{L.~Sun},
  \bibinfo{author}{C.~R. Johnson}, \bibinfo{author}{E.~Hendriks},
\newblock \bibinfo{title}{Matching canvas weave patterns from processing
  {X}-ray images of master paintings},
\newblock in: \bibinfo{booktitle}{IEEE International Conference on Acoustics,
  Speech and Signal Processing (ICASSP)}, \bibinfo{year}{2010}.
  \DOIprefix\doi{10.1109/ICASSP.2010.5495297}.
\bibitem[{Yang et~al.(2015)Yang, Lu, Brown, Daubechies, and Ying}]{Yang2015}
\bibinfo{author}{H.~Yang}, \bibinfo{author}{J.~Lu}, \bibinfo{author}{W.~P.
  Brown}, \bibinfo{author}{I.~Daubechies}, \bibinfo{author}{L.~Ying},
\newblock \bibinfo{title}{Quantitative canvas weave analysis using 2-{D}
  synchrosqueezed transforms: Application of time-frequency analysis to art
  investigation},
\newblock \bibinfo{journal}{IEEE Signal Processing Magazine}
  \bibinfo{volume}{32} (\bibinfo{year}{2015}) \bibinfo{pages}{55--63}.
\bibitem[{Lee(1980)}]{Lee80}
\bibinfo{author}{J.-S. Lee},
\newblock \bibinfo{title}{Digital image enhancement and noise filtering by use
  of local statistics},
\newblock \bibinfo{journal}{IEEE Transactions on Pattern Analysis and Machine
  Intelligence} \bibinfo{volume}{PAMI-2} (\bibinfo{year}{1980})
  \bibinfo{pages}{165--168}.
\bibitem[{Shorten and Khoshgoftaar(2019)}]{DA2019}
\bibinfo{author}{C.~Shorten}, \bibinfo{author}{T.~M. Khoshgoftaar},
\newblock \bibinfo{title}{A survey on image data augmentation for deep
  learning},
\newblock \bibinfo{journal}{Journal of big data} \bibinfo{volume}{6}
  (\bibinfo{year}{2019}) \bibinfo{pages}{1--48}.
\bibitem[{Ali et~al.(2022)Ali, Chuah, Talip, Mokhtar, and Shoaib}]{Ali2022}
\bibinfo{author}{R.~Ali}, \bibinfo{author}{J.~H. Chuah},
  \bibinfo{author}{M.~S.~A. Talip}, \bibinfo{author}{N.~Mokhtar},
  \bibinfo{author}{M.~A. Shoaib},
\newblock \bibinfo{title}{Crack segmentation network using additive attention
  gate---csn-ii},
\newblock \bibinfo{journal}{Engineering Applications of Artificial
  Intelligence} \bibinfo{volume}{114} (\bibinfo{year}{2022})
  \bibinfo{pages}{105130}.
\bibitem[{Zhang et~al.(2022)Zhang, Zhao, Chen, Zhou, Shi, and Yao}]{Zhang2022}
\bibinfo{author}{D.~Zhang}, \bibinfo{author}{J.~Zhao},
  \bibinfo{author}{J.~Chen}, \bibinfo{author}{Y.~Zhou},
  \bibinfo{author}{B.~Shi}, \bibinfo{author}{R.~Yao},
\newblock \bibinfo{title}{Edge-aware and spectral--spatial information
  aggregation network for multispectral image semantic segmentation},
\newblock \bibinfo{journal}{Engineering Applications of Artificial
  Intelligence} \bibinfo{volume}{114} (\bibinfo{year}{2022})
  \bibinfo{pages}{105070}.
\bibitem[{Yamanakkanavar and Lee(2022)}]{Yamanakkanavar2022}
\bibinfo{author}{N.~Yamanakkanavar}, \bibinfo{author}{B.~Lee},
\newblock \bibinfo{title}{Mf2-net: A multipath feature fusion network for
  medical image segmentation},
\newblock \bibinfo{journal}{Engineering Applications of Artificial
  Intelligence} \bibinfo{volume}{114} (\bibinfo{year}{2022})
  \bibinfo{pages}{105004}.
\bibitem[{Otsu(1979)}]{Otsu79}
\bibinfo{author}{N.~Otsu},
\newblock \bibinfo{title}{A threshold selection method from gray-level
  histograms},
\newblock \bibinfo{journal}{IEEE Transactions on Systems, Man, and Cybernetics}
  \bibinfo{volume}{9} (\bibinfo{year}{1979}) \bibinfo{pages}{62--66}.
\bibitem[{Kingma and Ba(2015)}]{Kingma15}
\bibinfo{author}{D.~P. Kingma}, \bibinfo{author}{J.~Ba},
\newblock \bibinfo{title}{{Adam: A Method for Stochastic Optimization}},
\newblock in: \bibinfo{booktitle}{Conf. for Learning Representations},
  \bibinfo{address}{San Diego, California, USA}, \bibinfo{year}{2015}.
  \URLprefix \url{http://arxiv.org/abs/1412.6980}.
  \href{http://arxiv.org/abs/1412.6980}{\tt arXiv:1412.6980}.
\bibitem[{de~Silva~y Vel{\'a}zquez(1632{\natexlab{a}})}]{P001167}
\bibinfo{author}{D.~R. de~Silva~y Vel{\'a}zquez}, \bibinfo{title}{The crucified
  {C}hrist}, \bibinfo{howpublished}{Museo Nacional del Prado (P001167)},
  \bibinfo{year}{1632}{\natexlab{a}}.
\bibitem[{de~Silva~y Vel{\'a}zquez(1632{\natexlab{b}})}]{P001196}
\bibinfo{author}{D.~R. de~Silva~y Vel{\'a}zquez}, \bibinfo{title}{Antonia de
  {I}pe{\~n}arrieta y {G}ald{\'o}s and her son, {L}uis},
  \bibinfo{howpublished}{Museo Nacional del Prado (P001196)},
  \bibinfo{year}{1632}{\natexlab{b}}.
\bibitem[{de~Silva~y Vel{\'a}zquez(1632{\natexlab{c}})}]{P001195}
\bibinfo{author}{D.~R. de~Silva~y Vel{\'a}zquez}, \bibinfo{title}{Diego del
  {C}orral y {A}rellano}, \bibinfo{howpublished}{Museo Nacional del Prado
  (P001195)}, \bibinfo{year}{1632}{\natexlab{c}}.
\bibitem[{Murillo(1965{\natexlab{a}})}]{P00998}
\bibinfo{author}{B.~E. Murillo}, \bibinfo{title}{The prodigal son taking leave
  of his home}, \bibinfo{howpublished}{Museo Nacional del Prado (P00998)},
  \bibinfo{year}{1660-1965}{\natexlab{a}}.
\bibitem[{Murillo(1965{\natexlab{b}})}]{P00999}
\bibinfo{author}{B.~E. Murillo}, \bibinfo{title}{The prodigal son squandering
  his inheritance}, \bibinfo{howpublished}{Museo Nacional del Prado (P00999)},
  \bibinfo{year}{1660-1965}{\natexlab{b}}.
\bibitem[{Ribera(1632)}]{P01114}
\bibinfo{author}{L.~S. Ribera, Jusepe~De}, \bibinfo{title}{{I}xion},
  \bibinfo{howpublished}{Museo Nacional del Prado (P01114)},
  \bibinfo{year}{1632}.

\end{thebibliography}


\end{document}